%% file: neurips_2025.tex
\documentclass{article}

\PassOptionsToPackage{numbers, compress}{natbib}

\usepackage[final]{neurips_2025}

\usepackage[utf8]{inputenc} 
\usepackage[T1]{fontenc}    
\usepackage{hyperref}       
\usepackage{url}            
\usepackage{booktabs}       
\usepackage{amsfonts}       
\usepackage{nicefrac}       
\usepackage{microtype}      
\usepackage{xcolor}         
\usepackage{xpatch}
\usepackage{graphicx}
\usepackage{subcaption}
\usepackage{multirow}
\usepackage{float}
\usepackage{tabularx}
\usepackage{adjustbox}
\usepackage{amsmath}
\usepackage[table]{xcolor}

\usepackage{amsmath}
\usepackage{amssymb}
\usepackage{mathtools}
\usepackage{amsthm}
\usepackage{wrapfig}
\usepackage{xcolor}
\usepackage{natbib}

\title{Handling Label Noise via Instance-Level Difficulty\\Modeling and Dynamic Optimization}

\author{%
  Kuan Zhang\textsuperscript{1},
  Chengliang Chai\textsuperscript{1,\thanks{Chengliang Chai is the corresponding author}},
  Jingzhe Xu\textsuperscript{1}, 
  Chi Zhang\textsuperscript{1}
  \\
  \textbf{Han Han}\textsuperscript{2},
  \textbf{Ye Yuan}\textsuperscript{1},
  \textbf{Guoren Wang}\textsuperscript{1},
  \textbf{Lei Cao}\textsuperscript{2}\\
  \textsuperscript{1}Beijing Institute of Technology\\
  \textsuperscript{2}University of Arizona\\
  \texttt{\{zhangkuan, ccl\}@bit.edu.cn} \\
}

\begin{document}

\maketitle

\begin{abstract}
Recent studies indicate that deep neural networks degrade in generalization performance under label noise. 
Existing methods focus on data selection or label correction, facing limitations such as high computational costs, heavy hyperparameter tuning process, and coarse-grained optimization.
To address these challenges, we propose a novel two-stage noisy learning framework that enables instance-level optimization through a dynamically weighted loss function, avoiding hyperparameter tuning. 
To obtain stable and accurate information about noise modeling, we introduce a simple yet effective metric, termed \textit{wrong event}, which dynamically models the cleanliness and difficulty of individual samples while maintaining computational costs. Our framework first collects \textit{wrong event} information and builds a strong base model. Then we perform noise-robust training on the base model, using a probabilistic model to handle the \textit{wrong event} information of samples. Experiments on six synthetic and real-world LNL benchmarks demonstrate our method surpasses state-of-the-art methods in performance, achieves a nearly 75\% reduction in storage and computational time, strongly improving model scalability. Our code is available at \url{https://github.com/iTheresaApocalypse/IDO}.
\end{abstract}

\input{intro}

\input{related_works}

\input{preliminary}

\input{framework}

\input{experiment}

\input{appf}

\section{Conclusion}
In this work, we introduce IDO, a novel framework that tackles noisy labels through fine-grained, instance-level optimization. Our approach is centered on the wrong event metric, a simple yet highly stable measure of sample cleanliness and difficulty that remains effective throughout training, even post-overfitting. By modeling this metric with a Beta Mixture Model, IDO uses a dynamic loss function that eliminates the need for manual hyperparameter tuning. Experiments confirm that IDO sets a new state-of-the-art in accuracy while being significantly more efficient and scalable than previous methods, offering a practical and robust solution for real-world noisy label learning.
\section*{Acknowledgments}
Chengliang Chai is supported by  National Key Research and Development Program of China (2024YFC3308200), NSFC (62472031), CCF-Baidu Open Fund (CCF-Baidu202402) and Huawei. 
Ye Yuan is supported by Beijing Natural Science Foundation (L241010), the National Key R\&D Program of China (Grant No.2022YFB2702100), and the NSFC (Grant Nos. 61932004, 62225203, U21A20516). 
Guoren Wang is supported by the NSFC (62427808, U2001211), the Liaoning Revitalization Talents Program (Grant No.XLYC2204005).
\bibliographystyle{IEEEtran}
\bibliography{Reference}


\appendix

\newpage

\begin{center}
    \hrule 

    \textbf{\large -Supplementary Material-} 
    \vspace{1mm}
    \hrule 
    
\end{center}
This supplementary material provides additional analysis and explanation of our paper, ``Handling Label Noise via Instance-Level Difficulty\\Modeling and Dynamic Optimization '', which were not included in the main manuscript due to page constraints. 

First of all, for the readers' better understanding, we describe the main content of the appendix. \textcolor{red}{Appendix A} details the sensitive analysis of two SOTA methods, including detailed parameter setting. \textcolor{red}{Appendix B} outlines the characteristics of each dataset and implementation details of learning rate, optimizer, baseline. \textcolor{red}{Appendix C} provides extensive theoretical and experimental analysis of wrong event for further analysis. In \textcolor{red}{Appendix D}, we illustrate the training time analysis.

\input{appa}

\input{appb}

\input{appc}

\input{appd}

\input{appe}


\newpage
\section*{NeurIPS Paper Checklist}

\begin{enumerate}

\item {\bf Claims}
    \item[] Question: Do the main claims made in the abstract and introduction accurately reflect the paper's contributions and scope?
    \item[] Answer: \answerYes{} 
    \item[] Justification: We believe the claims in our abstract and introduction accurately reflect the paper's contributions, including theoretical analysis and experimental results.

\item {\bf Limitations}
    \item[] Question: Does the paper discuss the limitations of the work performed by the authors?
    \item[] Answer: \answerYes{} 
    \item[] Justification: We've discussed the limitations in \autoref{appf}.

\item {\bf Theory assumptions and proofs}
    \item[] Question: For each theoretical result, does the paper provide the full set of assumptions and a complete (and correct) proof?
    \item[] Answer: \answerYes{} 
    \item[] Justification: We've provided clear descriptions and equations in the theoretical part, especially in \autoref{7} and \autoref{appc2}.

    \item {\bf Experimental result reproducibility}
    \item[] Question: Does the paper fully disclose all the information needed to reproduce the main experimental results of the paper to the extent that it affects the main claims and/or conclusions of the paper (regardless of whether the code and data are provided or not)?
    \item[] Answer: \answerYes{} 
    \item[] Justification: We've provided detailed information about the experiment setting in \autoref{2} and \autoref{appb}.

\item {\bf Open access to data and code}
    \item[] Question: Does the paper provide open access to the data and code, with sufficient instructions to faithfully reproduce the main experimental results, as described in supplemental material?
    \item[] Answer: \answerYes{} 
    \item[] Justification: We've provided our complete code in anonymous link mentioned in abstract. The datasets used in our paper are all open-source datasets which are easy to downloaNIPSd.

\item {\bf Experimental setting/details}
    \item[] Question: Does the paper specify all the training and test details (e.g., data splits, hyperparameters, how they were chosen, type of optimizer, etc.) necessary to understand the results?
    \item[] Answer: \answerYes{} 
    \item[] Justification: We've provided detailed information about the experiment setting in \autoref{2} and \autoref{appb}.

\item {\bf Experiment statistical significance}
    \item[] Question: Does the paper report error bars suitably and correctly defined or other appropriate information about the statistical significance of the experiments?
    \item[] Answer: \answerYes{} 
    \item[] Justification: We've mentioned in \autoref{2} that all experiments are conducted 5 times with random seed to confirm the statistical significance. The average results are reported in the paper.

\item {\bf Experiments compute resources}
    \item[] Question: For each experiment, does the paper provide sufficient information on the computer resources (type of compute workers, memory, time of execution) needed to reproduce the experiments?
    \item[] Answer: \answerYes{} 
    \item[] Justification:  We've mentioned in \autoref{2} that all experiments are conducted on a single Nvidia A100 80GB GPU.
    
\item {\bf Code of ethics}
    \item[] Question: Does the research conducted in the paper conform, in every respect, with the NeurIPS Code of Ethics \url{https://neurips.cc/public/EthicsGuidelines}?
    \item[] Answer: \answerYes{} 
    \item[] Justification: We confirm our paper conform with the NeurIPS Code of Ethics in every respect.

\item {\bf Broader impacts}
    \item[] Question: Does the paper discuss both potential positive societal impacts and negative societal impacts of the work performed?
    \item[] Answer: \answerYes{} 
    \item[] Justification: We've discussed the broader impact in \autoref{appf}.
    
\item {\bf Safeguards}
    \item[] Question: Does the paper describe safeguards that have been put in place for responsible release of data or models that have a high risk for misuse (e.g., pretrained language models, image generators, or scraped datasets)?
    \item[] Answer: \answerNA{} 
    \item[] Justification: The paper poses no such risks.

\item {\bf Licenses for existing assets}
    \item[] Question: Are the creators or original owners of assets (e.g., code, data, models), used in the paper, properly credited and are the license and terms of use explicitly mentioned and properly respected?
    \item[] Answer: \answerYes{} 
    \item[] Justification: Creators or original owners of assets (e.g., code, data, models), used in the paper are all properly credited and the license and terms of use are explicitly mentioned and properly respected.

\item {\bf New assets}
    \item[] Question: Are new assets introduced in the paper well documented and is the documentation provided alongside the assets?
    \item[] Answer: \answerNA{} 
    \item[] Justification: The paper does not release new assets.

\item {\bf Crowdsourcing and research with human subjects}
    \item[] Question: For crowdsourcing experiments and research with human subjects, does the paper include the full text of instructions given to participants and screenshots, if applicable, as well as details about compensation (if any)? 
    \item[] Answer: \answerNA{} 
    \item[] Justification: The paper does not involve crowdsourcing nor research with human subjects.

\item {\bf Institutional review board (IRB) approvals or equivalent for research with human subjects}
    \item[] Question: Does the paper describe potential risks incurred by study participants, whether such risks were disclosed to the subjects, and whether Institutional Review Board (IRB) approvals (or an equivalent approval/review based on the requirements of your country or institution) were obtained?
    \item[] Answer: \answerNA{} 
    \item[] Justification: The paper does not involve crowdsourcing nor research with human subjects.

\item {\bf Declaration of LLM usage}
    \item[] Question: Does the paper describe the usage of LLMs if it is an important, original, or non-standard component of the core methods in this research? Note that if the LLM is used only for writing, editing, or formatting purposes and does not impact the core methodology, scientific rigorousness, or originality of the research, declaration is not required.
    \item[] Answer: \answerNA{} 
    \item[] Justification: The core method development in this research does not involve LLMs as any important, original, or non-standard components.

\end{enumerate}

\end{document}

%% file: intro.tex
\section{Introduction}
\label{submission}
Curating large training datasets through web scraping \cite{Webvision01}, crowd-sourcing \cite{Yan02}, or pre-trained models \cite{BenitezQuiroz03} inevitably introduces label noise. It tends to degrade the performance of the trained deep neural network (DNN) model~\cite{zhang00} due to DNN memorizing and thus overfitting the noisy samples~\cite{MemoryEffect10}. 


To mitigate the negative impact of label noise, many works \cite{Coteaching04, O2U11,ELR05, Triple20,APL43, DISC14, TURN08, DEFT09, CLIPCleaner53} consider the patterns of the clean and noisy samples shown in the DNN training process, i.e., DNNs first learn clean patterns early in training, then start fitting noise later, and eventually overfit the noisy dataset. However, one key problem of this strategy is that the noisy samples tend to show behavior similar to the clean but difficult samples -- the clean examples that are close to the decision boundary of the model. To solve this problem, the most recent approaches~\cite{Triple20, DISC14} try to distinguish noisy examples from hard examples by introducing the hardness of the samples into the loss function. However, these methods have the following limitations.

\noindent \textbf{Limitations.} 
(1) These methods are {\it computationally intensive} due to the overhead for measuring the hardness and cleanness of the samples and grouping them into different categories, e.g., clean, noisy, and difficult \cite{UNICON33}; (2) these methods introduce {\it extra hyperparameters}, such as coefficients of loss terms, a cutoff threshold for grouping the samples, and a training epoch threshold to avoid the model fitting the noise, which are hard to tune \cite{APL43,TURN08}; (3) these methods always assign {\it identical coefficient} to all samples within the same category, neglecting the differences of individual samples in both cleanliness and difficulty \cite{DivideMix07, DISC14}.

%


\begin{wrapfigure}{r}{0.5\textwidth} 
    \centering
    \includegraphics[width=\linewidth]{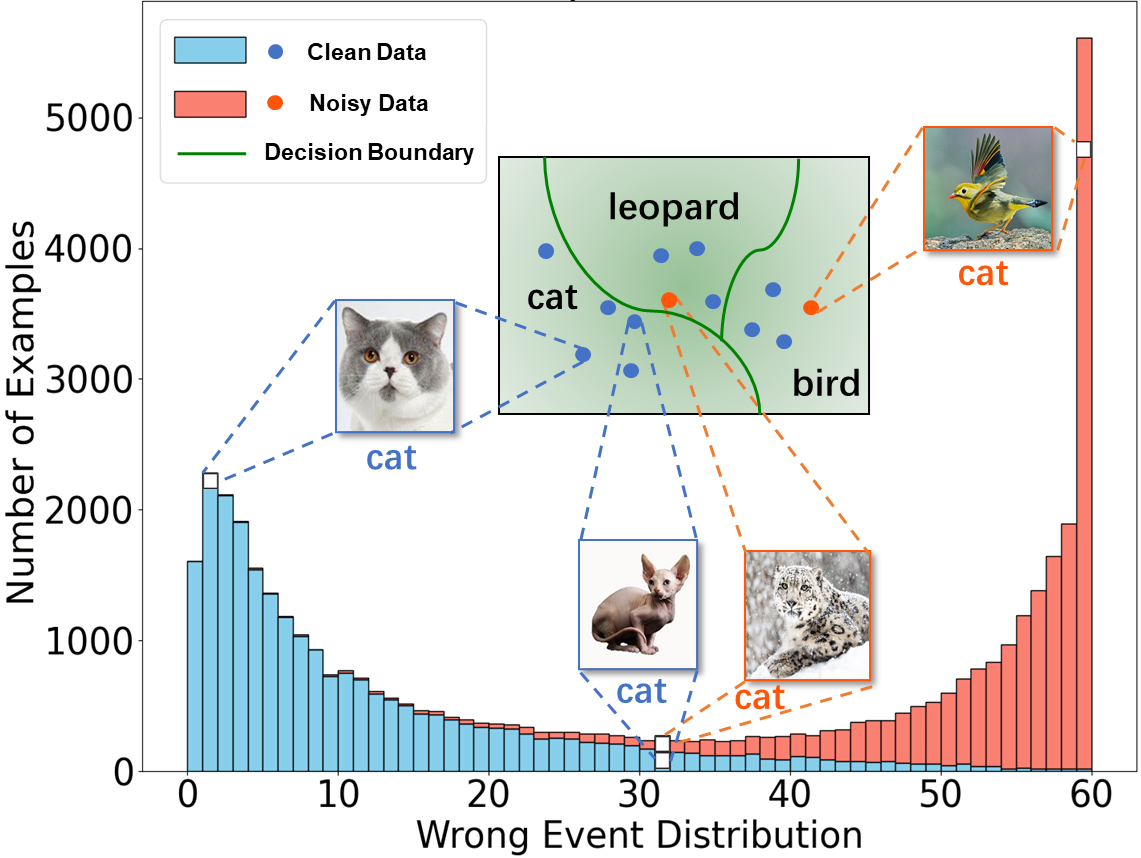}
    \caption{A bar chart illustrates the distribution of wrong events. In terms of sample cleanness, clean samples show lower wrong event, while noisy samples tend to have higher wrong event. In terms of sample difficulty, samples located at the extremes of the distribution represent easy samples, far from the decision boundaries (e.g., cat and bird in the figure). In contrast, samples located in the central depression of the distribution are often near the decision boundaries of similar classes, representing hard samples (e.g., cat and leopard in the figure).}
    \label{fig1}
\end{wrapfigure}

\noindent \textbf{Our Proposal.} To address these limitations, we propose  \texttt{IDO},  an \textbf{I}nstance-level \textbf{D}ifficulty Modeling and Dynamic \textbf{O}ptimization framework to achieve robust learning over noisy training data. Rather than relying on hyperparameters to regularize different terms in the loss function, \texttt{IDO} designs a dynamically weighted loss function that captures both the cleanliness and difficulty of each individual sample. This enables instance-level optimization without introducing any additional hyperparameters. 

\begin{figure*}[h]
\begin{center}
\includegraphics[width=1\textwidth]{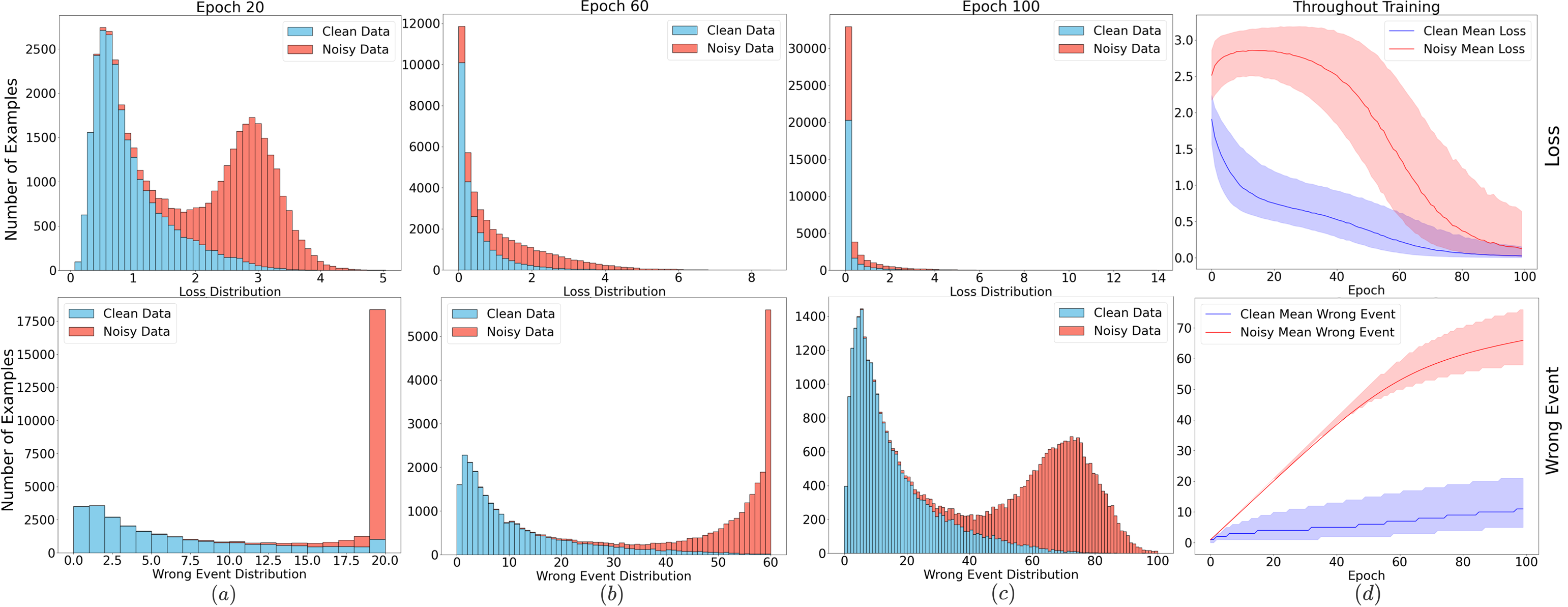}
\end{center}
\vspace{-1mm}
\caption{A comparison between wrong event and loss. The baseline model is ResNet-18 trained on CIFAR-10 with 40\% symmetric noise training for 100 epochs. We show the loss distributions (the first row) and wrong event distribution (the second row) during training. The four columns, (a) (b) (c) (d), represent the distributions at epoch 20, 60, 100 and the entire training phase. (a), (b), and (c) represent the early, middle, and late stages, respectively. In (d), the heavy lines represent the mean value and the shaded areas are the interquartile ranges. Since wrong events are monotonically increasing based on historical statistics instead of current model prediction, when model overfits the dataset, wrong event values for all samples do not change, rather than converging to zero as loss values typically do. As a result, wrong event can clearly separate noisy data and clean data in all training stages, even if the model fits noise.}
\vspace{-1mm}
\label{fig2}
\end{figure*}

\texttt{IDO} makes this possible by proposing a new noise-robust metric to replace the classical loss-based metric which is known to be unstable and thus ineffective in distinguishing hard and noisy samples~\cite{RoCL12}. This metric, called \textit{wrong event}, simply counts the frequency of mismatches between model predictions and the given labels, thus computationally efficient. Moreover, fitting a two-component beta mixture model to its distribution w.r.t. the training samples effectively measures the cleanliness and difficulty of each sample in a tuning-free manner (see Fig.\ref{fig1}). Replacing the hard-coded coefficients of the loss terms with the corresponding measures that dynamically update epoch by epoch
, \texttt{IDO} ensures accurate and robust modeling at any training stage (see Fig.\ref{fig2}) yet achieves high computational efficiency.

We summarize our main contributions as follows: (1) We propose \textit{wrong event}, a simple but effective metric that reliably separates noisy samples from clean ones at any stage of training, regardless of pretraining or not. We provide theoretical and experimental analysis to explain why \textit{wrong event} works well whether the model has fitted noise. (2) We propose \texttt{IDO} that trains a robust model with a dynamically weighted loss using \textit{wrong event} information where we are first to leverage probabilistic model to model data difficulty. (3) Our extensive experiments confirm that without hyperparameter tuning, \texttt{IDO} outperforms the state-of-the-art methods by an average accuracy of 1.6\% on synthetic and 0.7\% on real-world datasets for LNL tasks, with a 75\% reduction in computational time, and better scalability to larger models compared to previous methods.

%% file: related_works.tex
\section{Related works}
\textbf{Memorization \& Forgetting.} \cite{zhang00} observe that DNNs are sufficient for memorizing the entire dataset. \cite{MemoryEffect10} shows that DNNs follow the \textit{memorization effect}, i.e., clean samples and noisy samples exhibit distinct behaviors during the training process. Leveraging this observation, researchers have proposed other metrics to deal with noisy datasets. \cite{forgettingevent21} proposed \textit{forgetting event}, which counts how many times a model forgets a prediction of given labels. 
However, for easy noisy samples, the model tends to consistently predict ground truth labels rather than the given labels, resulting in the values of \textit{forgetting event} all being zero in early stage. This makes it difficult to distinguish clean and noisy samples. 
\cite{Latestop22} proposed \textit{first k-epoch learning} (\textit{fkl}), i.e., the time when a sample has been predicted to its given label for consecutive $k$ epochs for the first time. Samples with larger \textit{fkl} are more likely to be noisy. 
However, $k$ is a hyperparameter that varies on different datasets.
Although these metrics work well once the model training gets into the overfitting phase, they are less effective than loss in the early stage. Our proposed \textit{wrong event} metric demonstrates strong discriminative capabilities regardless of the stages of model training (see Fig.\ref{fig2}). More discussion about metrics is in Appendix~\ref{appc2}

\textbf{Learning with Noise Labels (LNL).} LNL problem has been extensively explored in recent research \cite{DISC14, DEFT09}. There are many variants of the LNL problem. Some methods \cite{soseleto37, MLC36} assume the existence of a small subset of clean data, which our method does not require. In our more challenging scenario, which has to distinguish noisy and difficult samples, recent studies leverage techniques such as co-training frameworks \cite{Coteaching04, DivideMix07, UNICON33}, k-means clustering \cite{rankmatch38}, and contrastive loss \cite{co-learning39, UNICON33}, which unfortunately incur high computational and memory costs. 
Dynamic loss functions can be considered as instance-level optimization. However, previous methods \cite{M-correction16, MLC36} mainly focus on modeling clean and noisy samples, while overlooking the challenge of modeling hard data, which our work targets.

\textbf{Pre-trained Models for LNL.} Pre-trained models are known for their strong generalization ability \cite{vit31, CLIP49}. Previous studies \cite{C2D29, SCL52} learn the representations to avoid label noise, by using self-supervised pre-trained models such as SimCLR \cite{SIMCLR51} and MoCo \cite{MoCO50}. 
Recent LNL methods \cite{TURN08, DEFT09, CLIPCleaner53}  leverage pre-trained vision models such as ResNet \cite{resnet32}, ViT \cite{vit31}, CLIP \cite{CLIP49} and ConvNeXt \cite{Convnext47} to improve effectiveness and efficiency of LNL. However, large pre-trained models restrict the applicability of complex modeling approaches. Therefore, they \cite{TURN08, DEFT09, CLIPCleaner53} primarily rely on simple strategies such as the small-loss criterion or similarity threshold to discard the potential noisy data from training. This loses the opportunity to explore the valuable information hidden in these noisy data. 
Our approach is much more scalable because of the lightweight \textit{wrong event}-based strategy. It is thus able to fully utilize both clean and noisy data, in turn achieving superior model performance.

%% file: preliminary.tex
\section{Key ideas: wrong event and instance-level optimization}

In this section, we first introduce the problem of LNL. Subsequently, we define the concept of \textit{wrong event} and present empirical evidence to demonstrate its robustness. Finally, we discuss how to fully leverage \textit{wrong event} in LNL.

\subsection{Noise label learning}

In the context of a $C$-class image classification problem 
, we denote a training dataset as $D_{\text{train}} = \{(x_i, \bar{y}_i)\}_{i=1}^N$, which consists of $N$ pairs of input images $x_i$ and their given labels $\bar{y}_i \in \{1, \dots, C\}$. In real-world scenarios, the given labels $\bar{y}_i$ may be corrupted due to various factors. We use $y_i$ to represent the ground truth label of $(x_i, \bar{y}_i)$, which remains inaccessible during training.

\subsection{Observation of wrong event}
\label{1}

We begin by formally defining the concept of \textit{wrong event}. Consider a model $f(\cdot)$ trained over $T$ epochs. For a given sample $(x_i, \bar{y}_i)$, the \textit{wrong event} is computed as follows:
\begin{equation}
\text{wrong event}_i = \sum_{t=1}^{T} \mathbb{I}(\arg\max_{i}f_t(x_i) \neq \bar{y}_i)
\end{equation}
where $\mathbb{I}(\cdot)$ denotes the indicator function. This metric quantifies the cumulative number of epochs in which the prediction of the model $f(x_i)$ disagrees with the given label $\bar{y}_i$.
We observe that \textit{wrong event} can reliably separate noisy samples from clean samples at any stage of training, even if the model has fitted noisy samples, as shown in Fig.\ref{fig2}. 
 
In addition to the robustness of \textit{wrong event}, i.e., the variance of \textit{wrong event} across consecutive epochs is much smaller than that of loss values, where the high variance tends to make loss ineffective~\cite{RoCL12} in separating hard and noisy samples, \textit{wrong event} effectively reflects both the cleanliness and difficulty of the training samples.  \textit{Easy samples}, which no matter clean or noisy, are located far from the decision boundary, exhibit consistent prediction behavior in consecutive epochs: clean (noisy) samples are consistently predicted correctly (incorrectly). As a result, these samples tend to occupy the extremes of the \textit{wrong event} distribution. In contrast, \textit{hard samples}, which lie close to the decision boundary, often experience fluctuating predictions, with their outcomes frequently flipping between similar classes. This instability causes hard samples to gather at the middle region of the \textit{wrong event} distribution, forming a low-lying trough. Therefore, the distribution of \textit{wrong event} offers valuable insights into both the cleanliness and the difficulty of each sample.

Our extensive theoretical and experimental analysis shows that \textit{wrong event} is indeed a more robust and informative metric than loss with less variance, regardless of pretraining or not (see Appendix~\ref{appc}) and provides more accurate measures than other metrics (see \autoref{ablation:metric}) in cleanliness and difficulty.

\subsection{Measuring cleanliness and difficulty}
\label{7}
To extract cleanliness and difficulty information from \textit{wrong event}, we employ a two-component probabilistic model to fit the distribution of \textit{wrong event}. In all the following equations, the symbol $w$ is used to denote the \textit{wrong event}. The probability density function (pdf) of a two-component mixture model on \textit{wrong event} is defined as:
\begin{equation}
p(w) = \sum_{k=1}^{2} m_{k} \cdot p(w|k)\label{mm}
\end{equation}
where $m_k$ are the mixing coefficients for the convex combination of each individual pdf $p(w | k)$. 

Existing methods commonly use the Gaussian mixture model (GMM) to distinguish clean samples from noisy ones \cite{DivideMix07, UNICON33}. However, in our case, the \textit{wrong event} of easy samples, especially easy noisy samples, often focuses on the distribution tails, leading to a monotonically increasing rather than unimodal distribution (see Figs.\ref{fig1},\ref{fig2}), making Gaussian distribution models inadequate for fitting. This phenomenon can be attributed to three main factors: \textbf{1)} Hard samples \cite{FocalLoss45, GHM46} often exhibit a long-tailed distribution in training; \textbf{2)} Curriculum learning \cite{CL44} suggests that models learn easy samples quickly but struggle with hard samples; \textbf{3)} Training typically stops before the model fully memorizes the dataset, leading to an accumulation of easy noisy samples at the tail of the noise distribution. 

To better model such distributions, we adopt a two-component beta mixture model (BMM), which is well suited to capture both symmetric, skewed, and monotonically increasing distributions due to the flexibility of the beta distribution \cite{bmm40}. We use an Expectation Maximization (EM) procedure \cite{M-correction16, DivideMix07} to fit the BMM to the observations. Unlike previous method \cite{M-correction16, DivideMix07}, specifically, we introduce two more variables to fully use statistical information to build sample cleanliness and difficulty: 
\begin{equation}
\tau_k(w) = p(k|w) = \frac{p(k)\cdot p(w|k)}{p(w)}
\end{equation}
which defines the posterior probability of a given $w$ having been generated by mixture component $k$,
\begin{equation}
\lambda_k(w) = F(w; \alpha_k, \beta_k) = \frac{\int_0^x t^{\alpha_k-1} (1-t)^{\beta_k-1} \, dt}{B(\alpha_k, \beta_k)}
\end{equation}
which defines the cumulative distribution function value of mixture component $k$ for a given $w$.

The fitting process yielded two beta distributions, $\mathcal{B}_1$ and $\mathcal{B}_2$, with means $\mu_1$ and $\mu_2$ (assuming that $\mu_1$ is smaller than $\mu_2$). We need to utilize $\mathcal{B}_1$ and $\mathcal{B}_2$ to measure the cleanliness and difficulty of the samples. 1) \textbf{Cleanliness.} We associate $\mathcal{B}_1$ with the clean distribution for lower mean, as it captures the characteristics of correctly labeled samples, while $\mathcal{B}_2$ is attributed to the noisy distribution, representing mislabeled samples. Thus, $\tau_1(\cdot)$ corresponds to the probability of a sample being drawn from the clean distribution, while $\tau_2(\cdot) = 1-\tau_1(\cdot)$ represents the probability of a sample originating from the noisy distribution. 2) \textbf{Difficulty.} In the clean distribution, samples with larger \textit{wrong event} values tend to be closer to the decision boundary. Conversely, in the noisy distribution, samples with smaller \textit{wrong event} values are closer to the decision boundary. We use posterior probabilities $\tau(\cdot)$ and the cumulative distribution function $\lambda(\cdot)$ to measure the difficulty level of the samples:
\begin{equation}
\epsilon(w)=\tau_1(w)\cdot\lambda_1(w) + \tau_2(w) \cdot (1-\lambda_2(w))\label{difficulty}
\end{equation}
where $\epsilon(w)$ represents the difficulty of a sample with the given \textit{wrong event}. $\tau(w)$ reflects the posterior probability in $w$ belonging to clean or noisy distribution. $\lambda(w)$ measures how extreme $w$ is within its assigned distribution. Low $\epsilon(w)$ indicates $w$ lies in a high-confidence, typical region of one component, while high $\epsilon(w)$ suggests $w$ resides in ambiguous regions. The larger (smaller) $\epsilon(w)$ is, the closer (farther) the sample is to the decision boundary, indicating a higher (lower) difficulty level of the sample. We analyze the bounds of $\epsilon(w)$:
\begin{equation}
0 \leq \epsilon(w) \leq \tau_1(w) + \tau_2(w)=1 \label{difficulty_analysis}
\end{equation}
Then we analyze its trend. When $ w \approx \text{min}\{w_i\}_{i=1}^N$, we have $\tau_1(w)=1,r_2(w)=0,\lambda_1(w)=0,\lambda_2(w)=0$, then $\epsilon(w)\approx 1\times0+0\times 1=0$, as does $w\approx \text{max}\{w_i\}_{i=1}^N$. When $ w \to \text{middle}$, we have $\tau_1(w)\approx 0.5,r_2(w)\approx 0.5,\lambda_1(w)\approx1,\lambda_2(w)\approx0$, then $  \epsilon(w) \approx 1 $.


Due to the heterogeneous distribution of \textit{wrong event} across different classes where simpler classes typically exhibit lower wrong event means compared to harder ones, fitting a single distribution to the entire dataset proves ineffective. To achieve more accurate modeling and maintain class balance, we establish $C$ BMM models to independently model the distribution of each class. For samples $i$ where $\bar{y_i}=c$, the posterior probabilities are derived from the corresponding BMM components $\mathcal{B}_1^c$ and $\mathcal{B}_2^c$, achieving more accurate and fine-grained modeling.

%% file: framework.tex
\section{Robust denoising framework}
\begin{figure*}[h]
\begin{center}
\includegraphics[width=1\textwidth]{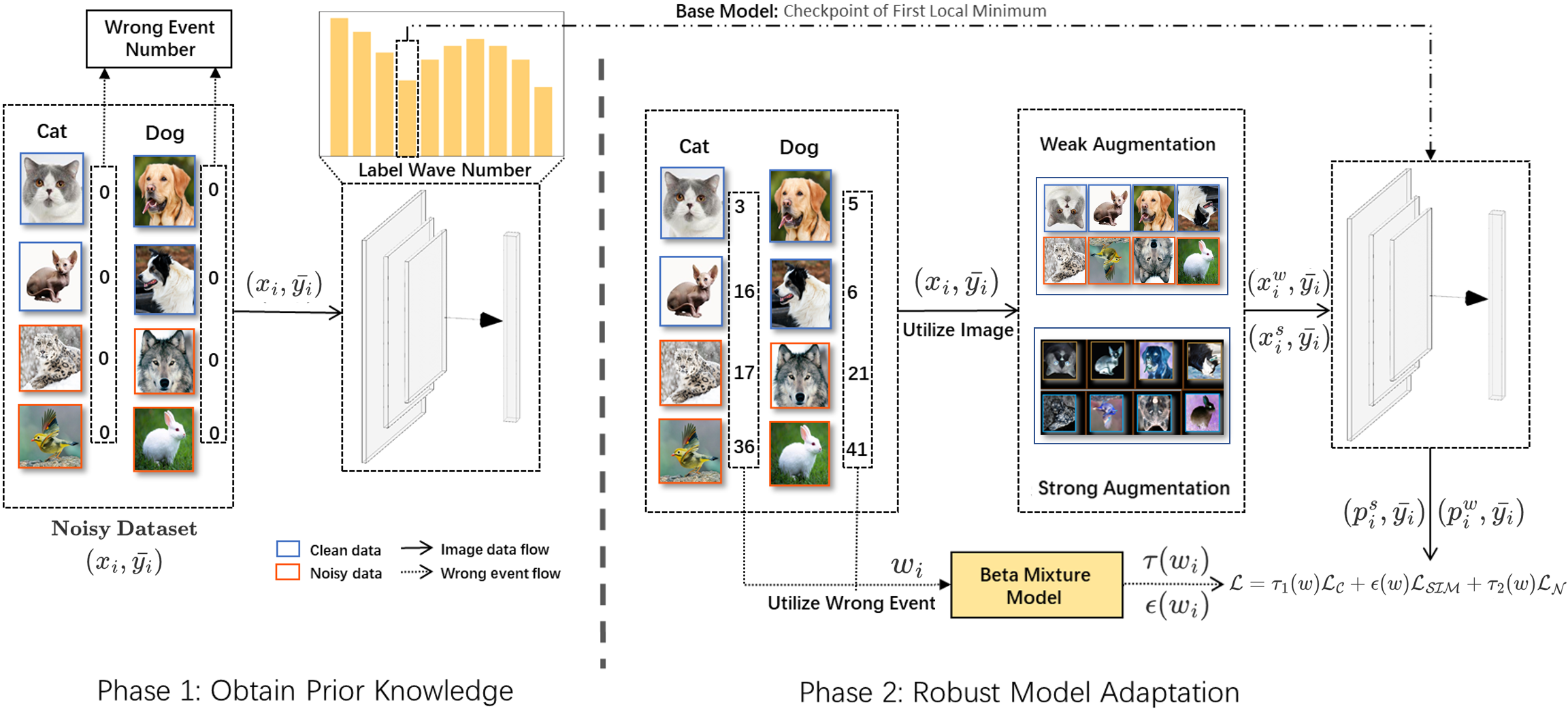}
\end{center}

\caption{Illustration of the proposed \texttt{IDO} framework. The training process is divided into two stages. In the first stage, prior knowledge, i.e., coarse distribution of wrong event, is obtained, and a base model that owns basic discrimination capability is captured. In the second stage, robust noise learning is performed. By using BMM, we obtain both cleanliness and difficulty information for individual samples, enabling instance-level dynamic optimization. The sample's wrong event information and the model's classification capability mutually benefit each other, leading to high improvement.}

\label{fig3}
\end{figure*}

We design a two-stage learning framework (see Fig.\ref{fig3}), which obtains the prior knowledge, i.e., \textit{wrong event} information and a competitive base model in the first stage, and then utilizes this prior knowledge to produce a noise-robust model in the second stage. 

\subsection{Two stage training}
\textbf{Stage 1: Obtain prior knowledge.}  The first phase has two main objectives. Firstly,  we adopted a typical training strategy, which involves training the model using the cross-entropy loss function for a certain period and collecting \textit{wrong event} information.
\begin{equation}
\mathcal{L}_{ce}(x_i,\bar{y})= -\bar{y} \ \text{log}(f(x_i))\label{pureCE}
\end{equation}
Secondly, we aim to capture a competitive base model during training to serve as the initial model for stage 2, thereby improving training quality. Recent research \cite{LabelWave15} introduced a metric called \textit{label wave}, which records the number of prediction changes per epoch. The model at the first local minimum of prediction changes exhibits strong competitiveness in LNL settings. In particular, this technique does not increase the computational cost of \texttt{IDO}. 
Notably, multiple fine-tuning approaches can be employed in stage 1, including Linear Probing (LP) or Full Fine Tuning (FFT). For smaller models such as ResNet-50, we recommend using FFT to achieve a more competitive base model and improve the accuracy of \textit{wrong event} estimation. For larger models such as ViT-B/16, we suggest using LP to accelerate training. The stage 1 process is illustrated in the left side of Fig.\ref{fig3}.

\textbf{Stage 2: Robust Model Adaptation.} The second phase of the algorithm aims to train a noise-robust model on top of the base model, incorporating the prior knowledge acquired in the first phase. The overall training objective of second phase is:
\begin{equation}
\mathcal{L} = \sum_{i=1}^N[\tau_1(w_i)\cdot \mathcal{L_{C}} + \epsilon(w_i)\cdot \mathcal{L_{SIM}} + \tau_2(w_i) \cdot \mathcal{L_{N}}]
\end{equation}
where $\tau_1(\cdot)$, $\tau_2(\cdot)$ are the posterior probabilities of $\mathcal{B}_1$, $\mathcal{B}_2$, and $\epsilon(\cdot)$ is the difficulty coefficient in Eq.\ref{difficulty}. We will give the formation of each loss function in detail in the following subsection.
After each epoch of robust training, we also update the \textit{wrong event} information for each sample. With more accurate model output, the clean and noisy distribution gradually separate. In this way, both the model and the \textit{wrong event} information become more accurate. The stage 2 process is illustrated in the right side of Fig.\ref{fig3}.

\subsection{Analysis of loss function}
For an instance $x$, two-view augmentation generates a weak view $x^w$ and a strong view $x^s$ \cite{AugDisc18, RRL17}. For clean distribution $\mathcal{B}_1$, we utilize the typical cross-entropy loss for both views, leading to the loss term $\mathcal{L_C}$:
\begin{equation}
\mathcal{L_{C}} = \mathcal{L}_{ce}(x_i^w,\bar{y}) + \mathcal{L}_{ce}(x_i^s,\bar{y})
\end{equation}
where $\mathcal{L}_{ce}$ is cross-entropy loss in Eq.\ref{pureCE}. For noisy distribution $\mathcal{B}_2$, we compute the linear average of the model outputs from the two views to obtain a robust pseudo-label to replace $\bar{y}$:
\begin{equation}
    f(x_i) = \frac{1}{2}f(x_i^w)+\frac{1}{2}f(x_i^s)\\
\end{equation}
which leads to the loss term $\mathcal{L_N}$, controlled by a confidence coefficient:
\begin{equation}
    \mathcal{L_N} = -c\cdot f(x_i)\ \text{log}(f(x_i)), c=\max(f(x_i))
\end{equation}

To optimize the difficult samples, we aim to enhance the feature extraction capability of the model w.r.t. these samples, as their labels and model outputs are often inaccurate. Specifically, we encourage the outputs of the two views to be as close as possible, and we use $\mathcal{L_{SIM}}$ to measure this similarity:
\begin{equation}
\mathcal{L_{SIM}} = (f(x_i^w)-f(x_i^s))^2
\end{equation}
$\mathcal{L_{SIM}}$ is the squared error (SE) between the two predictions. We use $\mathcal{L_{SIM}}$ to increase the loss weight for difficult samples. This encourages the model to focus more on learning from these samples, thereby achieving more robust feature extraction and classification capabilities.

%% file: experiment.tex
\section{Experiments}
\begin{table*}[h!]
\centering
\caption{Comparison with state-of-the-art LNL algorithms in test accuracy (\%) on CIFAR-100 and Tiny-ImageNet datasets.}
\scalebox{0.68}{
\begin{tabular}{lc|ccccc|cccc}
\toprule
\multirow{2}{*}{\textbf{Methods}} & \multirow{2}{*}{\textbf{Architecture}} & \multicolumn{5}{c|}{\textbf{CIFAR-100}} & \multicolumn{3}{c}{\textbf{Tiny-ImageNet}} \\
\cmidrule(lr){3-7} \cmidrule(lr){8-10}
 & & \textbf{Sym. 20\%} & \textbf{Sym. 40\%} & \textbf{Sym. 60\%} & \textbf{Asym. 40\%} & \textbf{Inst. 40\%} & \textbf{Sym. 20\%} & \textbf{Sym. 50\%} & \textbf{Inst. 40\%} \\
\midrule
Standard & RN-50 & 75.19±0.45 & 59.41±0.84 & 41.12±1.35 & 53.42±1.14 & 58.84±1.23 & 68.84±0.51 & 45.92±1.02 &53.72±1.18 \\
SCE & RN-50 & 75.21±0.51 & 69.37±0.65 & 55.25±0.71 & 67.04±0.81& 57.18±0.68 & 69.73±0.41 & 58.62±0.63 &67.36±0.89 \\
TURN & RN-50 &80.70±0.38 &78.68±0.44 & 73.36±0.53 & 69.49±0.69 & 69.82±0.93 &70.21±0.59 &67.52±0.81 &65.61±1.01 \\
ELR & RN-50 & 80.93±0.41 & 78.22±0.48 & 73.97±0.59 & 75.61±0.31 & 79.77±0.45 & 70.54±0.51 & 64.25±0.47 &75.62±0.68 \\
CoL & RN-50 & 83.15±0.35 & 81.76±0.40&79.22±0.59 &73.60±0.61  &80.18±0.49 &70.92±0.63 &67.38±0.78 &72.71±1.02 \\
DMix & RN-50 &84.27±0.26  &83.08±0.31 &80.69±0.36 &66.60±0.89  &81.37±0.40 &73.61±0.36 &71.47±0.42 &73.39±0.48 \\
UNICON & RN-50 &84.12±0.29  &\underline{83.20±0.28} &\underline{81.12±0.35} &76.77±0.39  &82.24±0.41 &75.61±0.47 &73.43±0.56 &74.71±0.49 \\
DISC & RN-50 &83.68±0.36 &82.25±0.40 &80.12±0.51 &\underline{77.12±0.44} &82.26±0.28  &75.73±0.42 &\underline{74.63±0.57} &\underline{75.79±0.26} \\
DeFT & CLIP-RN-50 & \underline{84.32±0.21} & 82.96±0.50 & 79.92±0.74 &69.92±0.81 & \underline{82.50±0.19} & \underline{76.81±0.19} & 73.92±0.37 & 74.16±0.35 \\

\rowcolor{gray!40}
IDO & RN-50 & \textbf{85.08±0.21} & \textbf{83.77±0.23} & \textbf{81.42±0.27} & \textbf{78.17±0.18}& \textbf{83.84±0.20} & \textbf{78.57±0.25} & \textbf{75.42±0.29} & \textbf{77.46±0.32} \\
\midrule
SCE & ViT-B & 91.26±0.20 & 90.38±0.29 & 87.42±0.37 &73.46±0.77 & 83.62±0.44 &87.51±0.27 &86.12±0.32 &83.22±0.51 \\
TURN & ViT-B &91.01±0.21 & 89.75±0.33 & 88.61±0.35 & \underline{85.12±0.25} & 84.67±0.30 &86.74±0.31 &75.32±0.47 &83.72±0.36 \\
ELR & ViT-B & 91.52±0.16 & 90.43±0.19 & \underline{89.74±0.20} &84.62±0.23 & \underline{91.42±0.16} & 87.22±0.29 &\underline{86.51±0.34} &\underline{87.92±0.26} \\
DeFT & CLIP-ViT-B & \underline{92.17±0.13} & \underline{91.23±0.17} & 89.42±0.25 & 72.96±0.81& 87.64±0.29 & \underline{89.04±0.22} & 72.93±0.62 &85.26±0.31 \\
\rowcolor{gray!40}
IDO & ViT-B & \textbf{92.67±0.09} & \textbf{92.36±0.13} & \textbf{91.45±0.13} & \textbf{89.65±0.12} & \textbf{92.24±0.15} & \textbf{91.25±0.15} & \textbf{90.21±0.14} &\textbf{90.32±0.19} \\
\bottomrule
\end{tabular}
}
\label{CombinedResults}
\end{table*}


\subsection{Experimental settings}\label{2}
\textbf{Synthetic Datasets.} We begin by evaluating the performance of \texttt{IDO} on three popular image classification benchmarks (CIFAR-10, CIFAR-100~\cite{cifar23} and Tiny-ImageNet~\cite{tinyimagenet24}) using synthetic datasets with varying types and ratios of noisy labels. 
%
For a noise transition matrix $T \in [0,1]^{K \times K}$,  $T_{ij}$ represents the probability of a ground-truth label $i$ being flipped to a corrupted label $\bar{y} = j$. Following previous works \cite{DISC14, DEFT09}, we introduce three common types of label noise: \textbf{1) Symmetric Noise}~\cite{Coteaching04, DivideMix07}: with noise rate $\eta$ and a $K$-class image classification task, we define $T_{ij} = \frac{\eta}{K}$ for $i \neq j$, where the true labels are replaced with random labels. 
\textbf{2) Asymmetric Noise}:  $T_{ij} = p(\bar{y} = j | y = i)$, which is designed to mimic the structure of real-world label noise. The labels are only replaced by similar classes. 
\textbf{3) Instance-dependent Noise}~\cite{IDN25}:  $T_{ij} = p(\bar{y} = j | y = i, x)$, which represents a more realistic scenario that considers the influence of instance $x$ in the label corruption process.

Following previous works~\cite{M-correction16, co-learning39}, we conduct experiments on CIFAR-100 with symmetric noise ratio $r \in \{0.2, 0.4, 0.6\}$, asymmetric noise ratio $r = 0.4$ and instance noise ratio $r = 0.4$, and on Tiny-ImageNet with symmetric noise ratio $r \in \{0.2, 0.5\}$ and instance noise ratio $r = 0.4$. \

\textbf{Real-World Datasets.} We further investigate the performance of \texttt{IDO} on three real-world noisy label datasets: \textbf{1) CIFAR-100N}~\cite{CIFAR100N26} ($r \approx 0.4$): A variant of CIFAR-100 with real-world human annotations collected from Amazon Mechanical Turk. \textbf{2) Clothing1M}~\cite{Clothing1m27} ($r \approx 0.385$): A large-scale dataset consisting of 1 million clothing images across 14 categories, collected from online shopping websites. \textbf{3) WebVision}~\cite{Webvision01} ($r \approx 0.2$): A dataset using 1,000 classes from ImageNet ILSVRC12, containing 2.4 million images crawled from Flickr and Google. Following previous works~\cite{DEFT09, TURN08}, we conduct experiments on the top 50 classes of the Google image subset.

\textbf{Architecture and baselines.} Using pre-trained models for noise-robust training can improve training quality~\cite{C2D29, efficient30, TURN08, DEFT09}. In our experiments, we primarily utilize two widely adopted pre-trained models: ViT-B/16  and ResNet-50. For ResNet-50, we compare \texttt{IDO} with state-of-the-art LNL algorithms, including one-stage methods SCE \cite{SCE41}, DivideMix (DMix) \cite{DivideMix07}, ELR \cite{ELR05}, Co-learning (CoL) \cite{co-learning39}, UNICON \cite{UNICON33}, DISC \cite{DISC14} and two-stage methods TURN \cite{TURN08} and DEFT \cite{DEFT09}. For ViT-B/16, we compare \texttt{IDO} with SCE, ELR, TURN and DEFT. Scalability analysis (see \autoref{scalability}) shows DMix and UNICON require 4× more time and 3× more memory, limiting scalability and making comprehensive comparisons on ViT-B/16 impractical. To verify the effectiveness of \texttt{IDO} on real-world datasets, we also compare with some state-of-the-art methods including LongReMix \cite{LongReMix34} and ProMix \cite{ProMix35}. The results of the baseline are reproduced by using the open-sourced code.

\textbf{Implementation.} Following setting in \cite{TURN08,DEFT09}. We run 5 epochs for stage one to obtain the prior knowledge about \textit{wrong event} for each sample, and run 10 epochs for stage two to fully robust train the pre-trained model. For two-stage baselines, we run 5 epochs for stage one and run 10 epochs for stage two to fully train the model. For one-stage baselines, we run 15 epochs to fully train the model. All experiment results are the averages of five random runs on a single A100 80G GPU. For detailed hyperparameter settings about the optimizer, dataset and baseline, please refer to \autoref{appb}.

\subsection{Performance for noisy label learning}\label{3}
\textbf{Overall performance on synthetic datasets.} \autoref{CombinedResults} demonstrates that \texttt{IDO} consistently outperforms competing methods in CIFAR100 and Tiny-ImageNet with an average accuracy of 1.4\%, achieving state-of-the-art performance on both large and small models. \texttt{IDO} outperforms DMix and UNICON especially in non-symmetric noise through its instance-level modeling. DeFT generally performs well in non-asymmetric noise, but faces challenges in resource-constrained scenarios due to its dependence on CLIP. DISC and ELR achieve good performance in non-symmetric noise but show limitation in symmetric noise, due to inadequate correction for symmetric noise. While existing methods struggle in some noise cases, \texttt{IDO} is able to handle different noise levels and types more effectively by difficulty modeling and dynamic optimization. Results of CIFAR-10 are in Appendix~\ref{appe4}.

\textbf{Overall performance on real-world datasets.} \autoref{realworld} demonstrates that \texttt{IDO} consistently outperforms competing methods in three real-world datasets with an average accuracy of 0.7\% on both large and small models. DeFT achieves promising results with dual-prompt technique but faces challenges in fine-grained classification tasks, such as Clothing1M. DMix and UNICON achieve competitive results in three datasets through dual-network subset partitioning, but with slower training efficiency and limitations in scalability. In challenging real-world noisy scenarios, loss regularization methods, i.e. SCE and ELR, do not perform well because they do not attempt to modify the noisy labels.

\begin{table*}[t]
\centering
\caption{Comparison with state-of-the-art LNL algorithms in test accuracy (\%) on CIFAR-100N, Clothing1M, WebVision dataset. The best results are highlighted in bold. Results marked with an asterisk (*) are from \cite{DEFT09}.}
\scalebox{0.75}{ 
\begin{tabular}{lccccc}
\toprule
\textbf{Method} & \textbf{Architecture} & \textbf{CIFAR-100N} & \textbf{Clothing1M-F} & \textbf{Clothing1M-R} &\textbf{WebVision} \\
\midrule
Standard &ResNet-50 &56.71 &67.52 &69.85 &77.84 \\
SCE &ResNet-50 &61.42 &69.65 &71.83 &78.35 \\
TURN &ResNet-50 &69.54 &68.32 &72.98 &79.28 \\
DeFT &CLIP-ResNet-50 &65.53 &70.46 &73.38 &76.62 \\
ELR &ResNet-50 &69.25 &70.52 &73.10 &80.16  \\
CoL &ResNet-50 &71.29 &72.04 &73.09 &80.96 \\
DMix &ResNet-50 &72.91 &72.23 &74.56&81.72 \\
UNICON &ResNet-50 &73.12 &\underline{72.44} &\underline{74.75} &82.48 \\
DISC &ResNet-50 &\underline{73.33} &72.15 &74.27  &\underline{82.69} \\
\rowcolor{gray!40}
IDO &ResNet-50& \textbf{73.58}& \textbf{72.65}& \textbf{74.85} &\textbf{82.92} \\
\midrule
SCE &VIT-16/B &76.45 &69.52 &73.13&83.25 \\
TURN &ViT-16/B &78.12 &70.31 &73.89& 83.44 \\
ELR  &ViT-16/B &79.04 &72.23 &\underline{74.39} &83.96 \\
UNICON* &CLIP-ViT-16/B & 77.76& 70.40&-& 84.63 \\
LongReMix* &CLIP-ViT-16/B & 74.09& 70.65&-& 84.96 \\
ProMix* &CLIP-ViT-16/B & 76.07& 70.79 & -& 84.55 \\
DeFT* &CLIP-ViT-16/B & \underline{79.12}& \underline{72.59}&-& \underline{85.21} \\
\rowcolor{gray!40}
IDO &ViT-16/B& \textbf{81.36}& \textbf{73.04} & \textbf{74.99} & \textbf{86.03} \\
\bottomrule
\end{tabular}
}
\label{realworld}
\end{table*}

\subsection{Further analysis}\label{4}
We also conduct extensive experiments to validate the superiority of the wrong event metric, the effectiveness, efficiency, and scalability of the proposed optimization framework. 

\textbf{Dynamic loss VS. hyperparameter-based methods.} Existing methods introduce extra hyperparameters, such as coefficients of loss terms, a cutoff threshold for grouping the samples and momentum coefficient, which are hard to tune. To demonstrate the advantages of dynamic loss, we conducted experiments comparing \texttt{IDO} with two classic algorithms, ELR and DMix. We adjusted the hyperparameters of them to compare with \texttt{IDO}. The results in \autoref{tab:ELRhyperparameters} and \autoref{tab:DMixhyperparameters} indicate that the value of the hyperparameters for ELR and DMix greatly affects the performance. Besides, the optimal parameters vary across different noise types and ratios, while \texttt{IDO} achieves the best results without hyperparameter tuning. For detailed information, please refer to \autoref{appa}.

\begin{wrapfigure}{r}{0.5\textwidth} 
    \centering
    \includegraphics[width=\linewidth]{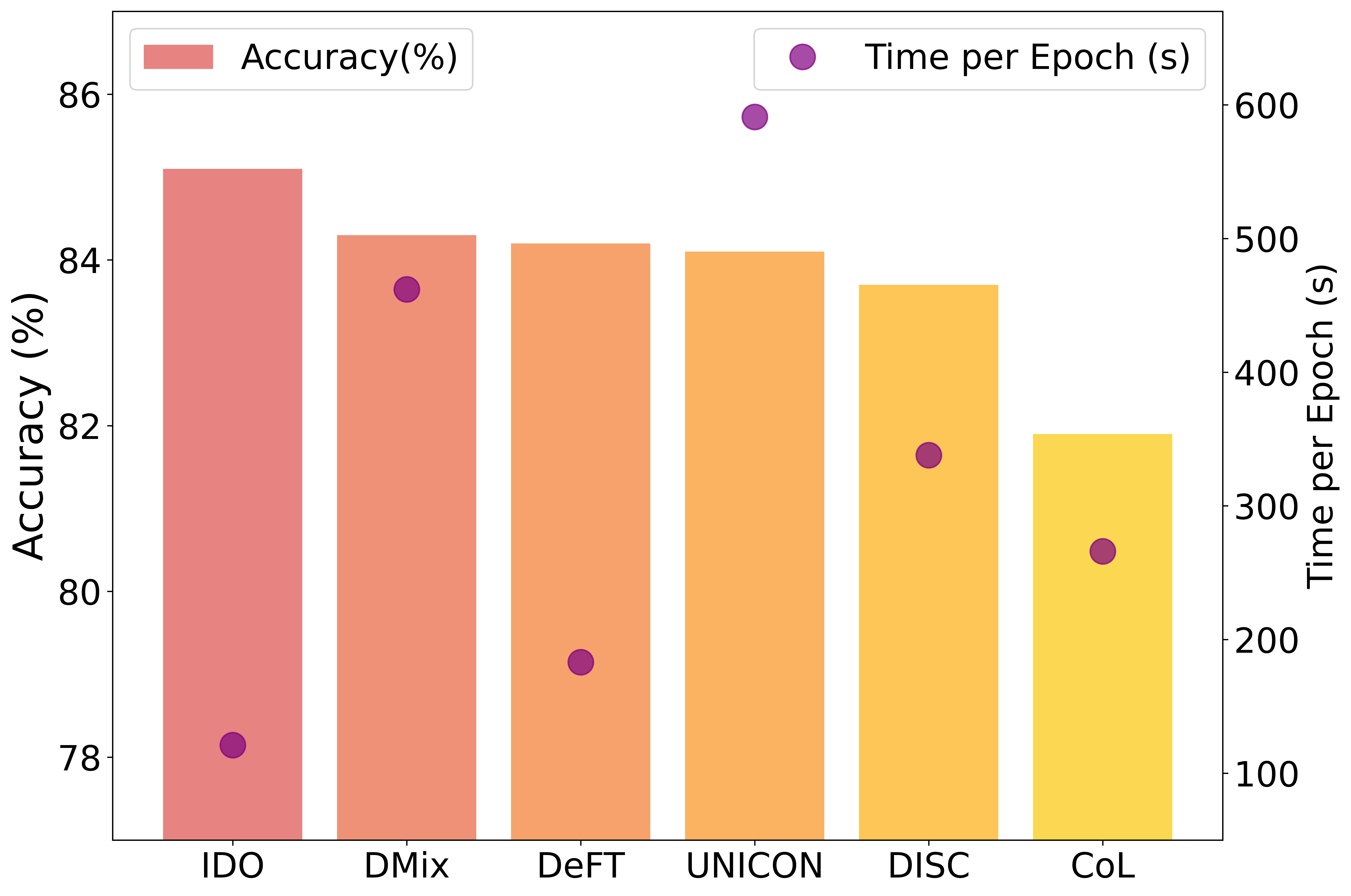}
    \caption{Comparison with state-of-the-art LNL algorithms in effectiveness and efficiency, using pre-trained ResNet50 on the A100 80GB GPU and CIFAR-100 dataset with 20\% symmetric noise.}
    \label{time}
\end{wrapfigure}

\textbf{Efficiency and scalability of \texttt{IDO}.} 
Existing methods have two drawbacks: high memory demand and low computing efficiency. For instance, DMix and UNICON train two networks simultaneously, which consumes a large amount of memory and limits scalability. Moreover, each sample requires 4 to 8 feed-forward computations, resulting in low efficiency. \texttt{IDO}, in contrast, only requires training a single network using \textit{wrong event} metric, which can be scaled to larger models. Additionally, each sample only needs two forward computations (two-views), achieving higher efficiency. Fig.\ref{time} shows that \texttt{IDO} achieves state-of-the-art performance while achieving high efficiency. \autoref{scalability} shows \texttt{IDO} is 4$\times$ faster and 3$\times$ less memory than DMix and UNICON on large models, with better performance on average. For detailed training time analysis, please refer to \autoref{appd}.

\textbf{\textit{Wrong event} VS. other metrics.} We adjust the BMM fitting distributions of different metrics to compare their performance in noise modeling. We change the duration of stage 1 to evaluate the metrics at different training stages. 
The results are presented in \autoref{ablation:metric}. Loss shows optimal performance in the early stage, but the optimal point is hard to identify; Even worse, the performance rapidly degrades in later stages due to noise overfitting. While the EMA-loss (Exponential Moving Average) mitigates noise overfitting in later training phases, it limits the capacity during the initial phase, creating a trade-off between early-stage and late-stage. The visualization about the clean sample selection ability of loss and wrong event is in Appendix~\ref{appc4}. Both \textit{forgetting event} (FE) and \textit{First k-epoch Learning} (FkL) show their noise modeling capabilities after the model begins to overfit the noise. However, these metrics need to accumulate over time to show differences, resulting in suboptimal performance in the early stages. Overall, \textit{wrong event} exhibits superior noise modeling capabilities in all training stages. The experiment results match our theoretical analysis in \autoref{1}, confirming the effectiveness of \textit{wrong event} in identifying noise. More analysis on metrics is in Appendix~\ref{appc3}~\ref{appc2}..

\begin{table*}[t]
    \centering
    \begin{minipage}{0.45\textwidth}
        \centering
        \caption{Comparison of methods scalability, running DMix, UNICON and \texttt{IDO} on CIFAR-100 Inst. 40\% noise with larger models on the A100 80GB GPU with a batch size of 64. We record the accuracy, average per epoch time (s) and the maximum GPU memory consumption (GB).}
        \scalebox{0.65}{ 
        \begin{tabular}{lccc|ccc}
            \toprule
            \textbf{Model} & \multicolumn{3}{c}{\textbf{ViT-B/16}} & \multicolumn{3}{c}{\textbf{ConvNeXt-B}} \\ 
            \cmidrule(lr){1-1} \cmidrule(lr){2-4} \cmidrule(lr){5-7}
            \textbf{Metric} & \textbf{Acc.} & \textbf{Time} & \textbf{Mem.}&\textbf{Acc.} & \textbf{Time} & \textbf{Mem.} \\
            \midrule
            DMix &89.5 &1596 &31.8 &88.5 &3197 &57.4 \\
            UNICON &90.5 &2628 &45.2&89.9 &4483 &78.9 \\
            \rowcolor{gray!40}
            IDO &\textbf{91.9} &\textbf{458} &\textbf{14.8}& \textbf{90.8}& \textbf{785} &\textbf{24.7}\\
            \bottomrule
        \end{tabular}
        }
        \label{scalability}
    \end{minipage}%
    \hfill
    \begin{minipage}{0.51\textwidth}
        \centering
        \caption{Comparison of noise modeling ability across 5 metrics at 4 training time in phase 1, using pre-trained ResNet50 on CIFAR-100. The training duration for phase 2 is set to 6 epochs. The '-' notation indicates cases where BMM fails to provide adequate fitting, due to all samples being assigned an identical metric value of 0.}
        \scalebox{0.66}{ 
        \begin{tabular}{lcccc|cccc}
            \toprule
            \textbf{Noise} & \multicolumn{4}{c}{\textbf{Sym. 60\%}} & \multicolumn{4}{c}{\textbf{Inst. 40\%}} \\ 
            \cmidrule(lr){1-1} \cmidrule(lr){2-5} \cmidrule(lr){6-9}
            \textbf{Epoch} & \textbf{1} & \textbf{2} & \textbf{4} & \textbf{8}  & \textbf{1}& \textbf{2} & \textbf{4} & \textbf{8}  \\
            \midrule
            Single Loss &\textbf{80.2}&\textbf{81.1} &\underline{79.5} &76.3  &\underline{78.2}&\underline{76.9} &74.6 &68.8 \\
            EMA Loss &79.9&\underline{80.9} &79.3 &77.9  &77.8&74.5 &73.6 &69.8  \\
            FkL &-&71.8 &73.2 &\underline{78.2}  &-&72.9 &\underline{75.2} &80.2 \\
            FE &-&65.8 &71.9 &75.8 &-&67.0 &73.2 &\textbf{81.7}  \\
            \rowcolor{gray!40}
            Wrong Event &\textbf{80.2}&\underline{80.9} &\textbf{80.4} &\textbf{79.6} &\textbf{82.7} &\textbf{83.6} &\textbf{82.6} &\underline{81.5}  \\
            \bottomrule
        \end{tabular}
        }
        \label{ablation:metric}
    \end{minipage}
\end{table*}

\begin{table*}[h!]
\centering
\caption{Ablation studies on the loss modules are conducted using ResNet50 under three different noise settings on CIFAR100.}
\scalebox{0.85}{ 
\begin{tabular}{lccccc}
\toprule
\multicolumn{3}{c}{\textbf{Loss Modules}} & \multicolumn{3}{c}{\textbf{CIFAR100}} \\
\cmidrule(lr){1-3}\cmidrule(lr){4-6}
\textbf{$\mathcal{L_C}$} & \textbf{$\mathcal{L_N}$} & \textbf{$\mathcal{L_{SIM}}$} & \textbf{Sym. 60\%} & \textbf{Asym. 40\%} & \textbf{Inst. 40\%} \\
\midrule
$\checkmark$ & & &75.1 &57.7 &70.5  \\
$\checkmark$ &$\checkmark$ & &\underline{79.5} &70.4 &77.3  \\
$\checkmark$ & &$\checkmark$ &78.2 &\underline{72.0} & \underline{80.3} \\
\rowcolor{gray!40} 
$\checkmark$ &$\checkmark$ &$\checkmark$ & \textbf{81.4}&\textbf{78.2} & \textbf{83.7}  \\
\bottomrule
\end{tabular}
}

\vspace{-5pt}
\label{ablation:Loss}
\end{table*}

\textbf{The effect of different loss terms.} Our loss function consists of three components: $\mathcal{L_C}$, $\mathcal{L_N}$ and $\mathcal{L_{SIM}}$,  modeling clean samples, noisy samples, and difficult samples, respectively. Our ablation experiments demonstrate that all three components contribute to improving the model performance (see Table \ref{ablation:Loss}). We observe that $\mathcal{L_N}$ achieves a higher improvement for symmetric noise, leveraging the accuracy of model output to improve classification performance, while $\mathcal{L_{SIM}}$ shows greater improvements for asymmetric and instance noise, improving the model's ability to learn from difficult samples by imposing consistency in predictions. More exploratory experiments are in \autoref{appe}.

\begin{table*}[h!]
\centering
\caption{Comparison of our dynamic weight $\epsilon(w)$ against several fixed-weighting methods on CIFAR-100 with different noise types}
\scalebox{0.8}{ 
\begin{tabular}{l|c|c|c}
\toprule
\textbf{Weighting Method} & \textbf{Sym. 60\%} & \textbf{Asym. 40\%} & \textbf{Inst. 40\%} \\
\midrule
Without $\epsilon=0$ & 79.5 & 70.4 & 77.3  \\
Fixed $\epsilon=0.25$ & 81.3 & 76.5 & 83.3  \\
Fixed $\epsilon=0.5$ & 80.8 & 77.5 & 82.4  \\
Fixed $\epsilon=1$ & 80.3 & 76.8 & 82.8  \\
Dynamic $\epsilon$ & \textbf{81.4} & \textbf{78.2} & \textbf{83.8}  \\
\bottomrule
\end{tabular}
}
\label{weighting}
\end{table*}

\textbf{The behavior of the difficulty-based weighting.} IDO relies more heavily on consistency learning under difficult conditions. We conducted new experiments comparing our dynamic weight $\epsilon(w)$ against several fixed-weighting methods (See Table \ref{weighting}). The result supports our core hypothesis, much like the principle behind Focal Loss \cite{FocalLoss45}, showing that a weighting term to expand the loss of difficult data is really necessary and better than treating all samples the same. We consider that for easy data, cross-entropy loss serves as a more direct and effective learning signal than consistency loss while consistency loss serves as a effective learning signal for difficult data. Dropping and fixing the difficulty coefficient both decrease the performance. More exploratory experiments are in \autoref{appe}.

%% file: appf.tex
\section{Limitations and broader impact}\label{appf}

\subsection{Limitations}
Wrong event becomes suboptimal when the noise ratio becomes extremely high, i.e., more than 90\%, because the distribution of wrong event will become extreme imbalance and the model struggles to converge. However, we do not suppose this will restrict the practical application ability of IDO, as real-world noise rates are often between 8.0\% and 38.5\% \cite{survey55}, far from extreme scenarios. However, designing a lightweight and highly scalable framework that remains competitive under extreme noise ratio is an important challenge for the community.

\subsection{Broader Impact}
Our research on label-noise learning contributes to building trustworthy machine learning systems, particularly in scenarios where high-quality annotations are costly or subjective (e.g., medical imaging diagnosis, social media content moderation). Consequently, as this method develops in effectiveness and scalability, the requirement for large-scale human-annotated data might decrease, potentially contributing to a rise in unemployment among data annotation professionals.

%% file: appa.tex
\section{Sensitivity analysis of parameter configurations in two SOTA methods}\label{appa}

In this section, we conduct comprehensive experiments on two advanced algorithms, ELR \cite{ELR05} and DMix \cite{DivideMix07}, to investigate the sensitivity of manually fixed parameters. Our analysis focuses on key hyperparameters including the subset division threshold, loss term weights, momentum averaging coefficients, among others.

\subsection{Analysis and experiments of ELR}
ELR algorithm incorporates two crucial hyperparameters: the regularization term weight $\lambda$ and the temporal ensembling momentum coefficient $\beta$. \autoref{tab:ELRhyperparameters} demonstrates the parameter sensitivity of ELR, showing that the performance is significantly influenced by parameter choices and comparison with the result of \texttt{IDO}. The parameter settings recommended for training from scratch in the original paper, where the authors recommend $\beta=0.7,\lambda=3$ for symmetric noise on CIFAR100 and $\beta=0.9,\lambda=7$ for asymmetric noise on CIFAR-100. However, parameter settings undergo significant shifts in pretrained settings. Optimal performance is achieved when $\beta\in[0.3,0.5],\lambda\in[3,5]$, demonstrating that optimal hyperparameters vary across different models and datasets, necessitating careful parameter tuning. Our proposed method \texttt{IDO} effectively addresses this challenge through dynamically weighted optimization.

\begin{table*}[h]
\centering
\caption{Hyperparameter sensitivity of ELR, utilizing pre-trained ResNet50 on the CIFAR100 dataset with three noise settings. The default values are set as $\lambda = 3$ and $\beta = 0.7$. Each time, one parameter is perturbed while others are set to default. We consider the value ranges of $\lambda \in \{0, 1, 3, 5, 7, 10\}$ and $\beta \in \{0.1, 0.3, 0.5, 0.7, 0.9, 0.99\}$, following original paper \cite{ELR05}.}
\scalebox{0.8}{
\begin{tabular}{ccccccc|cccccc|c}
\toprule
Hyperparameter & \multicolumn{6}{c}{regularization term weight $\lambda$} & \multicolumn{6}{c}{momentum coefficient $\beta$} & \texttt{IDO}  \\
\cmidrule(lr){1-1}  \cmidrule(lr){2-7} \cmidrule(lr){8-13}
Noise Dataset & 0 & 1 & 3 & 5 & 7 & 10 & 0.1 & 0.3 & 0.5 & 0.7 & 0.9 & 0.99 \\
\midrule
CIFAR-100 Sym. 60\% &41.1 &41.7 &73.9 &\textbf{75.9} &46.4 &1.0 &12.2 &75.9 &\textbf{77.1} &73.9 &51.3 &42.9 &\textbf{81.0} \\
CIFAR-100 Inst. 40\% &58.8 &56.0 &79.7 &\textbf{82.9} &69.5 &25.0 &49.9 &\textbf{82.9} &82.6 &79.7 &67.3 &59.4 & \textbf{83.6}\\
CIFAR-100N &61.0 &61.6 &\textbf{69.2} &68.8 &62.7 &9.1 &56.1 &70.7 &\textbf{71.8} &69.2 &65.3 &62.1 &\textbf{73.6}\\
\bottomrule
\end{tabular}
}
\label{tab:ELRhyperparameters}
\end{table*}

\subsection{Analysis and experiments of DivideMix}
DivideMix algorithm incorporates three crucial hyperparameters: the unsupervised loss term weight $\lambda_u$, GMM threshold $\tau$ for subset partitioning and beta distribution parameter $\alpha$ of Mixup \cite{mixup48}. \autoref{tab:DMixhyperparameters} demonstrates the parameter sensitivity of DivideMix, showing that the performance is significantly influenced by parameter choices and comparison with \texttt{IDO}'s result. The parameter settings recommended for training from scratch in the original paper, where the authors recommend $\lambda_u=25, \tau\in[0.5,0.6],\alpha=4$ for symmetric noise on CIFAR100 and $\lambda_u=0, \tau=0.5,\alpha=4$ for asymmetric noise on CIFAR-100. However, parameter settings undergo significant shifts in pretrained settings. Optimal performance is achieved when $\lambda_u\in[50,150], \tau\in[0.1, 0.5]$ shows that optimal hyperparameters vary between different models and data sets, which requires careful parameter tuning. Our proposed method \texttt{IDO} effectively addresses this challenge through dynamically weighted optimization.

\begin{table*}[h]
\centering
\caption{Hyperparameter sensitivity of DivideMix, utilizing pre-trained ResNet50 on the CIFAR100 dataset with three noise settings. The default values are set as $\lambda_u = 50$, $\tau = 0.5$ and $\alpha=4$. Each time, one parameter is perturbed while others are set to default. We consider the value ranges of $\lambda_u \in \{0, 25, 50, 150\}$, $\tau \in \{0.1, 0.3, 0.5, 0.7, 0.9\}$ and $\alpha \in \{0.5, 1, 2,4\}$ following original paper \cite{DivideMix07}.} 
\scalebox{0.752}{
\begin{tabular}{ccccc|ccccc|cccc|c}
\toprule
Hyperparameter & \multicolumn{4}{c}{$\mathcal{L_U}$ term weight $\lambda_u$} & \multicolumn{5}{c}{GMM threshold  $\tau$}  & \multicolumn{4}{c}{Mixup parameter $\alpha$}   & \texttt{IDO}  \\
\cmidrule(lr){1-1}  \cmidrule(lr){2-5} \cmidrule(lr){6-10} \cmidrule(lr){11-14}
Noise Dataset &0 & 25 & 50 & 150 & 0.1 & 0.3 & 0.5 & 0.7 & 0.9 & 0.5 & 1 & 2 & 4 &  \\
\midrule
CIFAR-100 Sym. 60\% &80.7 &79.3 &79.9 &\textbf{80.8} &\textbf{81.1} &80.5 &79.6 &67.4 &29.8 &\textbf{80.9} &79.7 &79.2 &78.0   &\textbf{81.0} \\
CIFAR-100 Inst. 40\% &79.7 &79.9 &\textbf{81.9} &81.1 &\textbf{80.3} &79.3 &79.7 &78.1 &70.5 &79.8 &81.1 &\textbf{82.3}&81.3 & \textbf{83.6}\\
CIFAR-100 Asym. 40\% &63.2 &63.6 &66.6 &\textbf{67.9} &63.8 &65.3 &\textbf{66.6} &65.9 &57.6 &60.3 &62.1 &66.4 &\textbf{66.6} & \textbf{77.8}\\
\bottomrule
\end{tabular}
}
\label{tab:DMixhyperparameters}
\end{table*}

%% file: appb.tex
\section{Additional implementation details}\label{appb}
\textbf{Regarding the hyperparameters of the model.} In our experiments, we primarily utilized pre-trained ResNet-50, ViT-16/B and ConvNeXt-B models, both of which were obtained by calling the PyTorch timm library. For the pre-training hyperparameters of the model, we adhered to the settings used in prior work \cite{TURN08, DEFT09}, as detailed in \autoref{tab:optimizer_config}. One-stage baselines follow the setting in stage 2.
\begin{table*}[h!]
\centering
\caption{Optimizer configurations for different models and stages.}
\scalebox{0.8}{
\begin{tabular}{ccccccc}
\toprule
\textbf{Phase} & \multicolumn{3}{c}{\textbf{Stage 1}} & \multicolumn{3}{c}{\textbf{Stage 2}} \\
\cmidrule(lr){1-1} \cmidrule(lr){2-4} \cmidrule(lr){5-7}
\textbf{Configuration} & \textbf{ViT-B/16} & \textbf{ResNet-50} & \textbf{ConvNeXt-B}& \textbf{ViT-B/16} & \textbf{ResNet-50} & \textbf{ConvNeXt-B}\\
\midrule
\textbf{Optimizer} & SGD & AdamW & AdamW & SGD & AdamW & AdamW \\
\textbf{Learning Rate} & $1 \times 10^{-2}$ & $1 \times 10^{-3}$ & $1 \times 10^{-4}$ & $1 \times 10^{-2}$ & $1 \times 10^{-3}$ & $1 \times 10^{-4}$\\
\textbf{Weight Decay} & $1 \times 10^{-5}$ & $1 \times 10^{-5}$ & $1 \times 10^{-4}$ & $1 \times 10^{-5}$ & $1 \times 10^{-5}$ & $1 \times 10^{-4}$\\
\textbf{Scheduler}  & No & No  & No & Cosine & Cosine & Cosine \\
\textbf{Strategies} & LP & FFT & LP & FFT & FFT & FFT\\
\bottomrule
\end{tabular}
}

\label{tab:optimizer_config}
\end{table*}

\textbf{Regarding the details of the datasets.} The model quality was determined by the accuracy on the test set at the last epoch. For synthetic datasets, we added noise to the entire training set and used the test set to evaluate performance. CIFAR100 consists of 100 classes, with 50,000 training images and 10,000 test images, and we set the batch size to 128. Tiny-ImageNet contains 200 classes, with 100,000 training images and 10,000 test images, and we also set the batch size to 128. 

For real-world datasets, we only used the noisy datasets and did not utilize the clean subsets provided by the datasets. Clothing1M is a class-imbalanced dataset, and we sample class-balanced subsets each time, with a batch size of 64 and 1,000 iterations. Clothing1M-F means using \textbf{f}ixed subset during training, used in \cite{DEFT09, TURN08}. Clothing1M-R means sampling \textbf{r}andom subset per epoch, used in \cite{DivideMix07, DISC14}. In Clothing1M-R, we calculate wrong event for the subset every 100 iterations. CIFAR100N is identical to CIFAR100 except for its real-world noisy labels. For WebVision, we used the first 50 classes, which include 66000 training images and 2,500 test images, and we set the batch size to 128.

\textbf{Regarding the hyperparameters of the baselines.} The baseline results were generated under consistent experimental settings using publicly available code. The hyperparameters were configured according to the recommendations outlined in the original papers.

SCE \cite{SCE41} is a robust noise-tolerant loss function with two hyperparameters, $\alpha$ and $\beta$. The original paper suggests that a large $\alpha$ may lead to overfitting, while a small $\alpha$ can mitigate overfitting but slow down convergence. Therefore, the authors recommend using a small $\alpha$ and a large $\beta$ to replace existing loss functions. Since pre-trained models converge faster, we adopted the configuration proposed in the original paper to suppress overfitting to noise, setting $\alpha=0.1, \beta=1$.

ELR \cite{ELR05} introduces a regularization term to the loss function, leveraging the correct predictions of noisy samples during early learning to achieve noise-robust learning with two hyperparameters, $\lambda$ and $\beta$. The authors recommend hyperparameter settings of CIFAR100, Clothing1M and WebVision, with $\lambda=3, \beta=0.7$. We applied the same settings across these datasets. For CIFAR100N and Tiny-ImageNet, we adopted the same configuration as used for CIFAR100. 

DivideMix \cite{DivideMix07} is a co-training-based label correction method that introduces five key hyperparameters: number of MixMatch views $M$, temperature of sharpen labels $T$, mixup parameter $\alpha$, GMM threshold $\tau$ and unsupervised loss term coefficient $\lambda_u$. We followed the original paper's settings, for most experiments, setting $M = 2$, $T = 0.5$, $\alpha = 4$, $\tau = 0.5$, and  $\lambda_u = 50$. 

Co-learning \cite{co-learning39} is a method combining supervised learning and self-supervised learning, with $L=L_{sup}+L_{int}+L_{str}$. All the coefficients of loss terms are set to $1$ following the original paper.

UNICON \cite{UNICON33} is a co-training-based label correction method that combines supervised learning, semi-supervised learning and contrastive learning. There are seven main parameters mentioned: unsupervised loss coefficient $\lambda_u$, regularization coefficient $\lambda_r$, contrastive loss coefficient $\lambda_C$, filter coefficient $\tau$, adjustment threshold $d_\mu$ and mixup coefficient $\alpha$. Following the original paper, we set $\lambda_C=0.025, \lambda_u=30, \lambda_r=1, d_\mu=0.7, \tau=5$ and $\alpha=4$.

DISC \cite{DISC14} is a method which dynamically selects and corrects dataset. Four main hyperparameters are mentioned: warm-up period $T_0$, coefficient of the hard-set loss $\lambda_h$, momentum coefficient $\lambda$ and positive offset value $\sigma$. Following the pre-trained model setting in the original paper, we set $T_0=1$, $\lambda=0.7$, $\lambda_h=0.2$ and $\sigma=0.3$.

TURN \cite{TURN08} is a method which firstly uses noise-robust loss function to obtain a clean set with LP, and FFT the model on the clean set. The main hyperparameter is the GMM threshold $\tau$. Following the original paper, we set $\tau=0.6$ in all experiments.

DeFT \cite{DEFT09} is a CLIP-based method which used dual prompts on CLIP to obtain a clean set with parameter-efficient fine-tuning (PEFT), and FFT the pre-trained downstream model on the clean set. The heavy reliance on CLIP significantly limits the method's applicability in computational resource-constrained scenarios.

\textbf{Regarding the hyperpapameters of BMM.} Since IDO relies on no priors, all experiments default that $B_1(1,2), B_2(2,1)$. The EM algorithm will update the parameters to the appropriate values.

%% file: appc.tex
\section{More analysis of wrong event}\label{appc}
\begin{figure*}[t]
\begin{center}
\includegraphics[width=0.98\textwidth]{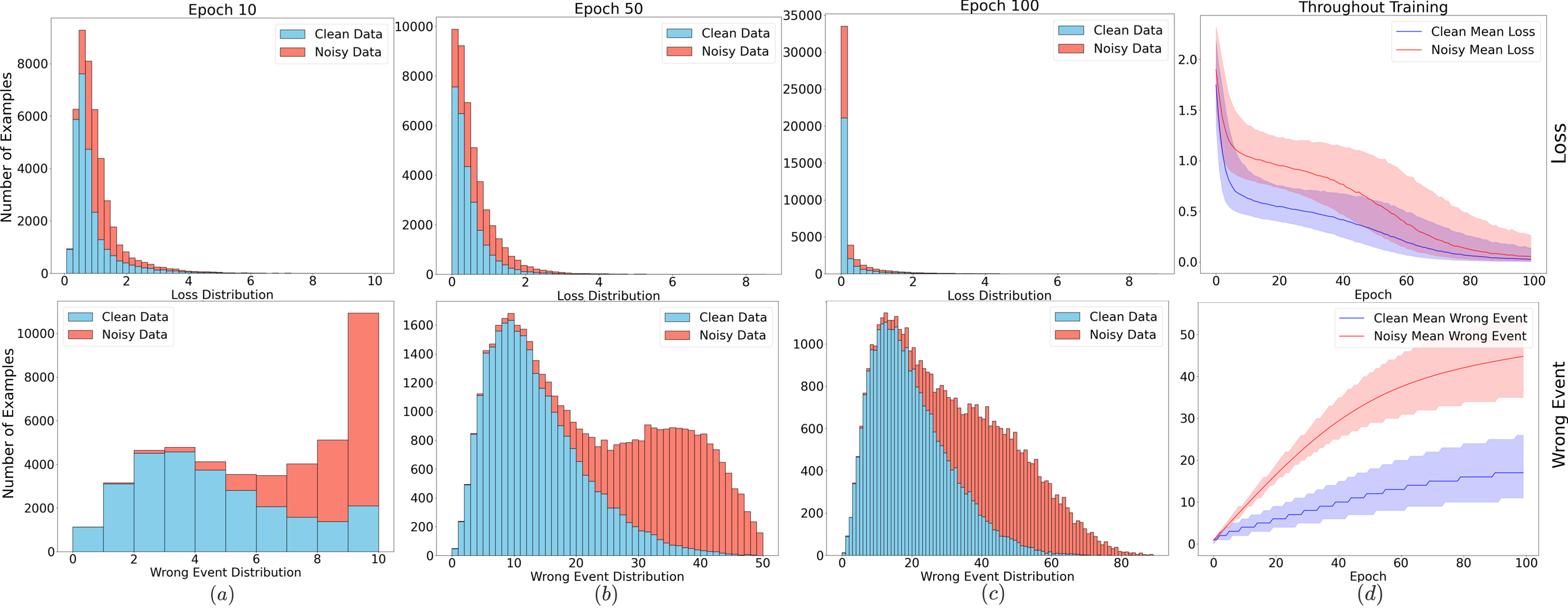}
\end{center}
\vspace{-1mm}
\caption{The baseline model is ResNet-18 trained on CIFAR-10 in 40\% asymmetric noise for 100 epochs. We show the loss distributions (the first row) and wrong event distribution (the second row) during training. The four columns, (a) (b) (c) (d), represent the distributions at epoch 10, 50, 100 and the entire training phase. In (d), the heavy lines represent the median value and the shaded areas are the interquartile ranges, respectively. It is clear that wrong event can clearly separate noisy data and clean data, even if the model fits noise. }
\vspace{-1mm}
\label{exp-asym}
\end{figure*}

\begin{figure*}[t]
\begin{center}
\includegraphics[width=0.98\textwidth]{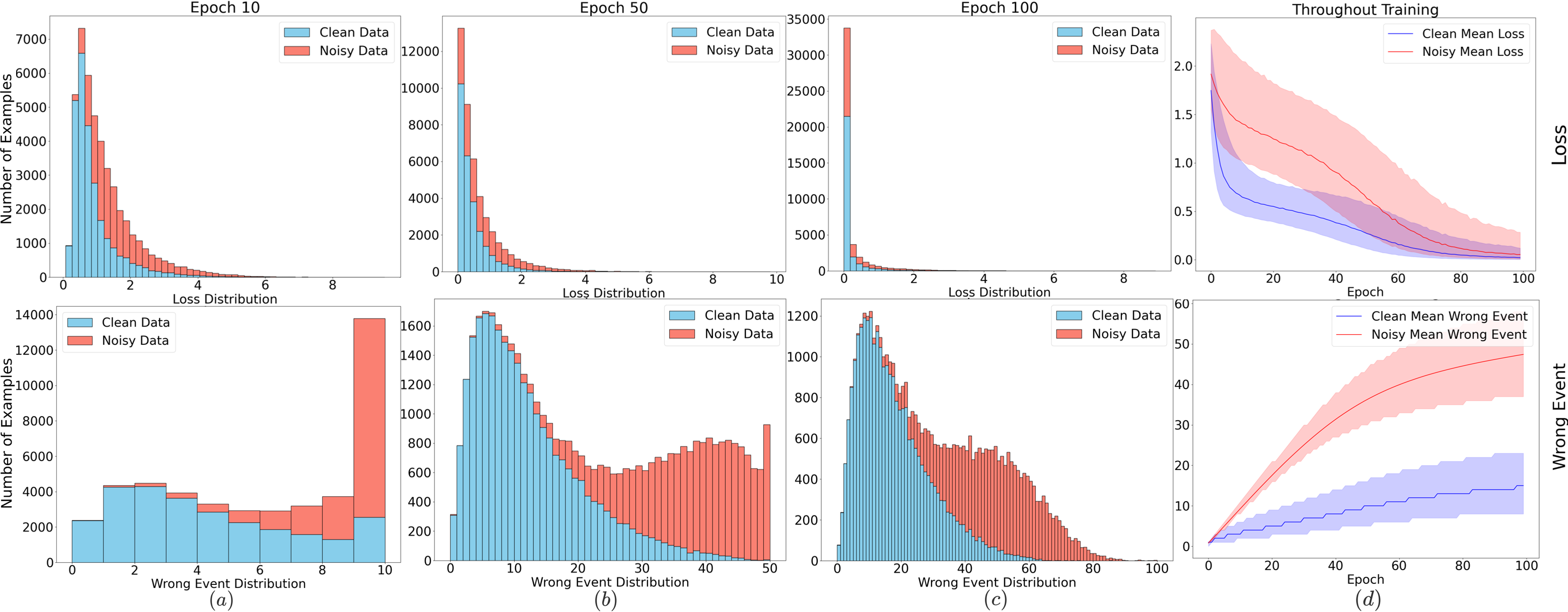}
\end{center}
\vspace{-1mm}
\caption{The baseline model is ResNet-18 trained on CIFAR-10 with 50\% instance noise for 100 epochs. We show the loss distributions (the first row) and wrong event distribution (the second row) during training. The four columns, (a) (b) (c) (d), represent the distributions at epoch 10, 50, 100 and the entire training phase. In (d), the heavy lines represent the median value and the shaded areas are the interquartile ranges, respectively. It is clear that wrong event can clearly separate noisy data and clean data, even if the model fits noise. }
\vspace{-1mm}
\label{exp-idn}
\end{figure*}
\subsection{More empirical observations of wrong event}\label{appc1}
We show a comparison between wrong event and loss under symmetric noise (see Fig.\ref{fig1}). We also conduct experiments under instance noise, and asymmetric noise to verify that \textit{wrong event} outperforms the loss across all types of noise. Fig.\ref{exp-asym} and Fig.\ref{exp-idn} show a comparison between wrong event and loss under asymmetric noise and instance noise, respectively. The experiment results reveal that in more challenging noisy environments, the model tends to rapidly fit the loss of noisy samples, resulting in compromised accuracy and stability of the provided modeling information. In contrast, wrong event metric maintains its capability to deliver robust modeling information about sample cleanliness and difficulty.

\subsection{Analysis on the change rate between wrong event and loss}\label{appc3}

We provide an analysis about the change rate between wrong event and loss. The loss change rate is \(\frac{\delta_{i}^{T+1} - \delta_{i}^{T}}{\delta_{i}^{T}} \in [-\infty, +\infty]\) in every epoch \(T\) with unbounded limits and frequently unpredictable mutations. A usually seen phenomenon named \textit{forgetting event} \cite{forgettingevent21} (last epoch right, this epoch wrong) can cause the probability of given label to plummet from a high probability (e.g.~\(\sim 0.9\)) to a small probability (e.g.~\(\sim 0.1\)), causing the loss to skyrocket and the relative change to become enormous. This can happen randomly and repeatedly for all samples.

However, for wrong event, the change rate needs more discussion. Firstly, we have \(w_{i}^{T+1} = w_{i}^{T} + \delta_{i}^{T+1}, \delta_{i}^{T+1} \in \{0, 1\}\) and \(w_{i}^{T} \in [0, \ldots, T]\), meaning the wrong event value is monotonically increasing. The change rate is \(\frac{\delta_{i}^{T+1}}{w_{i}^{T}}\). We analyze this in two distinct cases:

\textbf{The ``0 to 1'' Transition:} When a sample is misclassified for the very first time, \(w_{i}^{T}\) transitions from 0 to 1. At this point, the relative change \(\frac{1}{0}\) is not defined. This is a singular, one-time event for any given sample, marking its initial entry into the set of ``ever-misclassified'' samples.

\textbf{All Subsequent Changes (\(w_{i}^{T} \geq 1\)):} For every subsequent misclassification, the denominator \(w_{i}^{T}\) is at least 1. The relative change is \(\frac{\delta_{i}^{T+1}}{w_{i}^{T}}\). \textbf{This value is strictly bounded within the range [0,1]} far smaller than the change rate of loss. More importantly, as the model continues to misclassify a sample, \(w_{i}\) increases, causing the maximum possible relative change (\(\frac{1}{w_{i}^{T}}\)) to decrease. The metric is therefore self-stabilizing.

The relative change of wrong event is well-behaved, bounded, and self-stabilizing after a single initial event. Conversely, the relative change of loss is unbounded and perpetually susceptible to explosive volatility. This fundamental difference in their rate of change is why wrong event serves as a far more reliable and stable signal for modeling sample characteristics in noisy environments.

\begin{figure}[h]
    \centering
    \begin{subfigure}{0.49\textwidth}
        \centering
        \includegraphics[width=\linewidth]{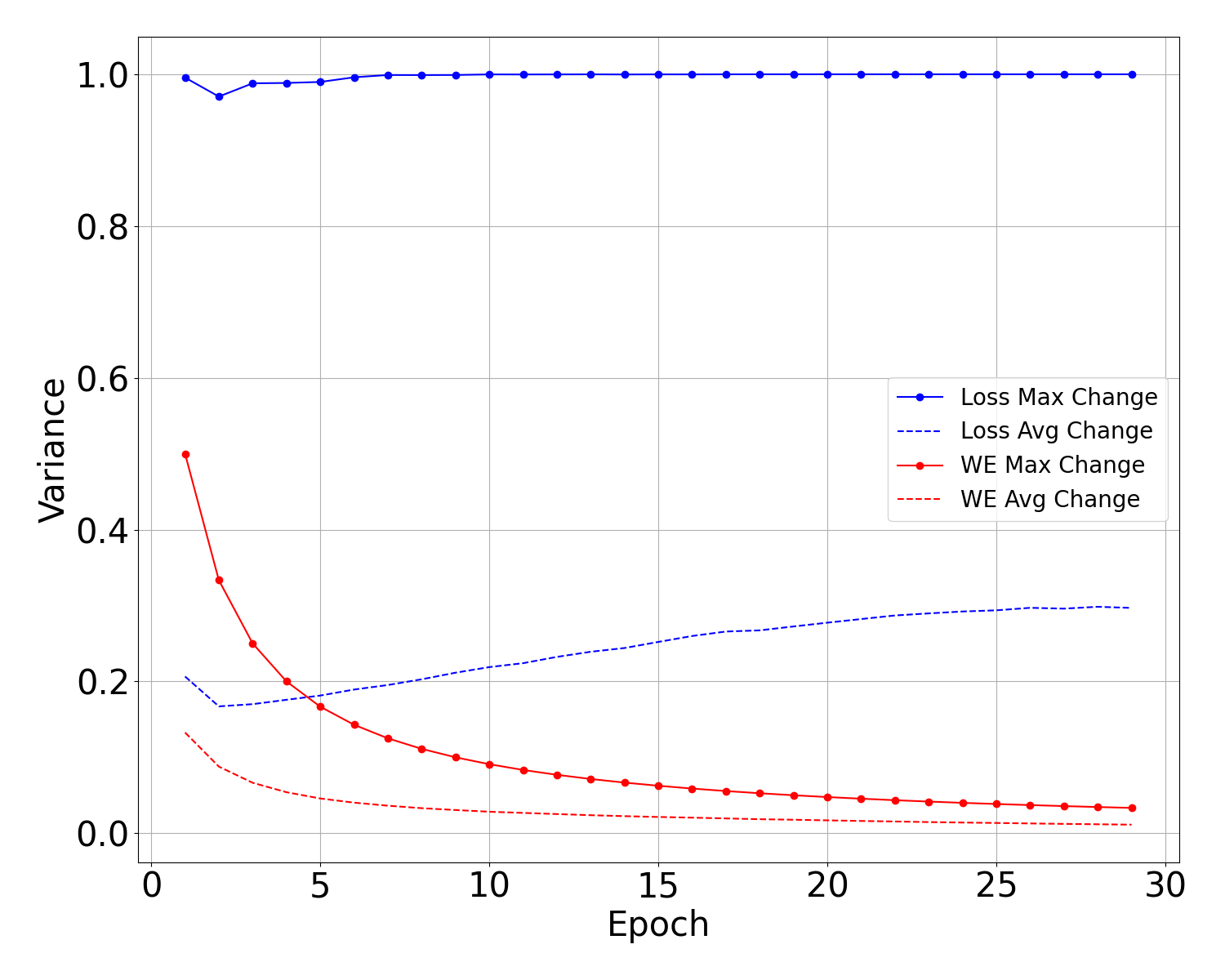}
        \caption{Variance over epochs on pre-training setting}
        \label{fig-v1}
    \end{subfigure}
    \hfill
    \begin{subfigure}{0.49\textwidth}
        \centering
        \includegraphics[width=\linewidth]{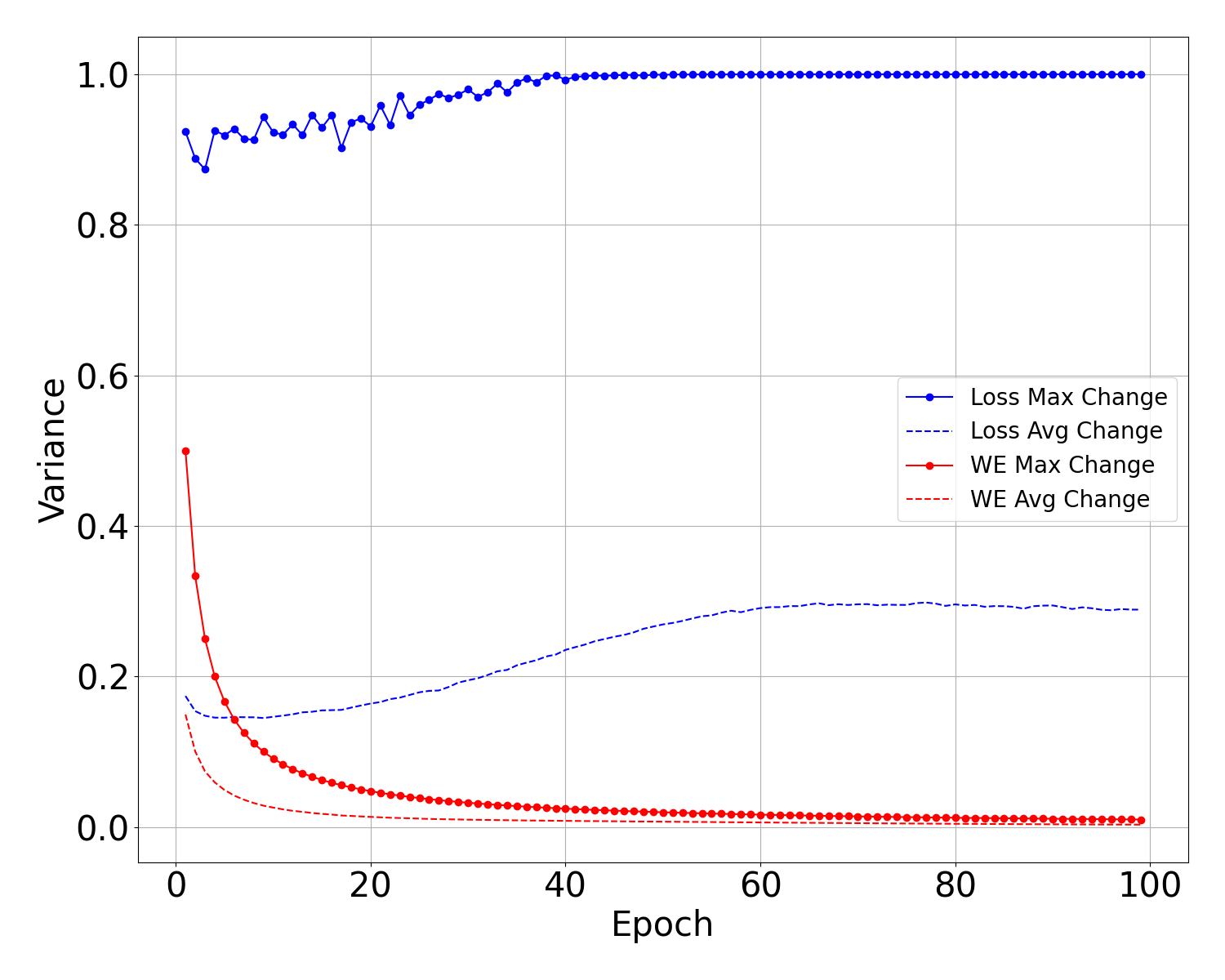}
        \caption{Variance over epochs on random initialized setting}
        \label{fig-v2}
    \end{subfigure}
    \caption{Variance comparison between loss and wrong event (WE). We test random pre-trained ResNet-50 on CIFAR-100 Inst. 40\% noise (left) and random initialized ResNet-18 on CIFAR-10 Inst. 40\% noise (right).}
    \label{fig-variance}
\end{figure}

This analysis provides a strong theoretical basis for the empirical stability we observed in Figure \ref{fig-variance}. In experiments, the values of loss and wrong are not in the same range. We normalize metric values and track the range of value changes during training $\delta_i(l_i^t) = |l^t_i - l^{t-1}_i|,\delta_i(w_i^t)=|w_i^t-w_i^{t-1}|$ and $\text{Max Change}=\mathop{\text{argmax}}\limits_{(x_i, \bar{y_i})\in \mathcal{D}}(\delta_i), \text{Avg Change}=\frac{1}{N}\sum_{i=1}^n\delta_i$. The result is in \autoref{fig-variance}. Loss variance is near the size of the dataset, confirming that loss has no bound. Wrong event variance decreases over training. Besides, due to the mutation of loss, the mean change is larger than wrong event. Experiment results confirm the theoretical analysis regardless of pre-training or not.

\subsection{More discussion about different metrics}\label{appc2}
Researchers have proposed some metrics trying to replace loss because loss is easy to fit noise, such as forgetting event \cite{forgettingevent21}, fluctuation event \cite{fluctuateevent54}, first k-epoch learning \cite{Latestop22}. All of them are based on forgetting. In all the following equations, the symbol $\hat{y_i}^t$ is used to denote the model prediction of sample $(x_i,\bar{y_i})$ at epoch $t$ . Forgetting event (FE) is defined as
\[
\text{forgetting event}_i=\sum_{t=1}^T (\bar{y_i}=\hat{y_i}^{t-1}) \wedge (\bar{y_i}\neq\hat{y_i^t})
\]
which counts how many times a model forgets a prediction of given labels. Fluctuation event is defined as  
\[
\text{fluctuation event}_i=(\bar{y_i}=\hat{y_i^{t_1}}) \wedge (\bar{y_i}\neq\hat{y_i^{t_2}})
\]
where $t_1\leq t_2$. First k-epoch learning (fkl) is defined as
\[
\text{fkl}_i= \mathop{\text{argmin}} \limits_{t^*\in t}[(\bar{y_i}=\hat{y_i}^{t^*}) \wedge \cdots \wedge (\bar{y_i}=\hat{y_i}^{t^*-k+1}) = 1]
\]
which records the epoch when a sample has been predicted to its given label for consecutive k epochs for the first time. 

\begin{figure}[t]
  \centering
  \begin{minipage}[t]{0.495\textwidth}
    \centering
    \includegraphics[width=0.8\linewidth]{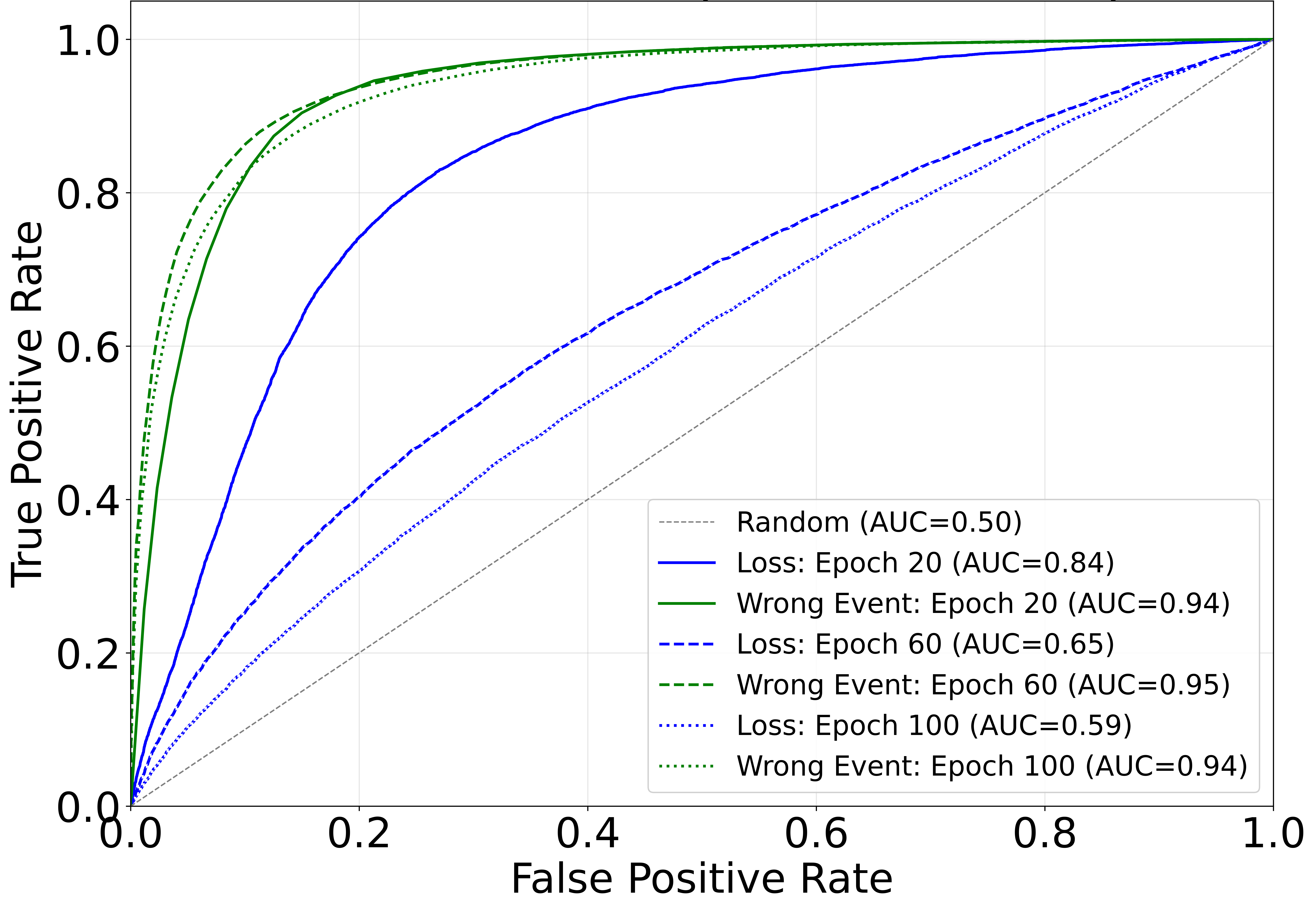} 
  \end{minipage}%
  \hfill 
  \begin{minipage}[t]{0.495\textwidth}
    \centering
    \includegraphics[width=0.8\linewidth]{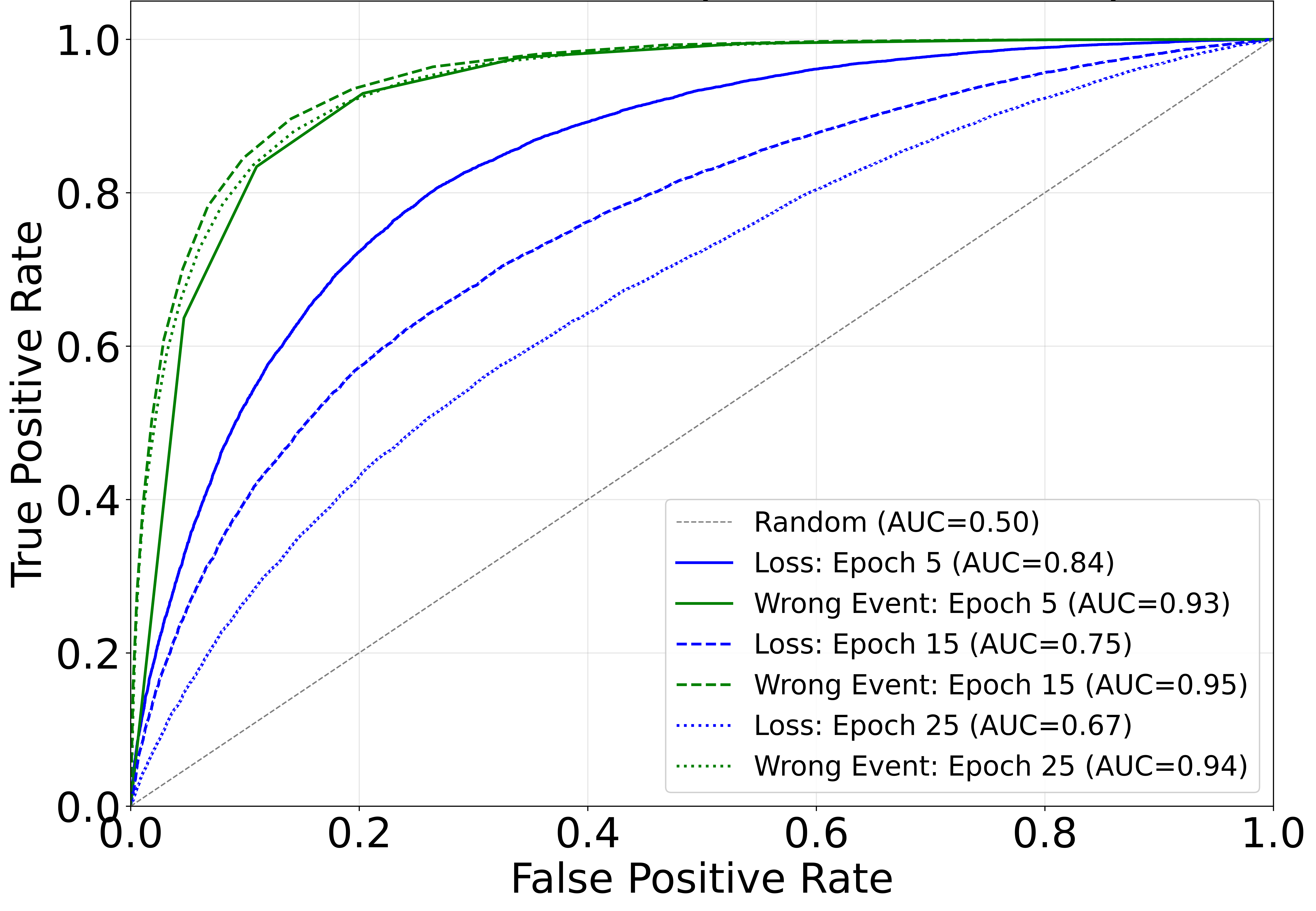} 
  \end{minipage}
\caption{The AUC-ROC curves of loss and wrong event. The experiment is conducted with random initialized ResNet-18 on CIFAR-10 (left) and pre-trained ResNet-50 on CIFAR-100 (right) under Inst.40\% noise. The curves show the dynamic selection curves over epochs. We visualize the ability of clean sample selection in early, midterm and later training stages. The curves show that wrong event has a larger AUC value than loss and show good performance during the whole training phase, which represents a better ability of clean sample selection.}
\label{auc-plot}
\end{figure}

We group all three metrics into forgetting-based metrics (FBM). All metrics extract the statistical information to assist noise modeling, but are still different. Notably, FBM needs time to remember the given label, because the model needs time to fit the given label, especially when the label is noisy while wrong event needn't. This difference makes \textbf{FBM spend pretty longer time accumulating statistics.} Considering easy noisy data for example, in early stage, model learns correctly and ground-truth of easy noisy data is consistently predicted, so the model cannot remember noisy labels. Easy noisy data's wrong event would increase rapidly, but FBM would be constant value. Thus, FBM cannot distinguish easy noisy data in early stage, needing more training time and thus the scalability is limited, while wrong event has good modeling ability during the training phase. \autoref{ablation:metric} also shows that loss-based metrics perform well in early stage but decrease over training, and FBM such as FE and fkl perform poorly in early stage but increase over training, needing more training time. While wrong event perform well during the entire training phase, showing effectiveness and efficiency.

\subsection{Selection ability between wrong event and loss}\label{appc4}
\autoref{ablation:metric} shows that wrong event outperforms loss in test accuracy. To further compare their clean sample selection ability, we calculate F-score along with precision and recall, AUC along with accuracy and precision under pre-training and random initialized settings. \autoref{scr-f} and \autoref{pre-f} show the F-score, \autoref{random-auc} and \autoref{pre-auc} show the AUC value and \autoref{auc-plot} shows the AUC-ROC curves during training. All results indicate that wrong event is better than loss in selecting clean samples. Wrong event can clearly model the noise during the whole training phase, while loss only performs well in early stage. In conclusion, wrong event is a simple but effective metric to select clean samples, which can capture noise well even if the model has fitted the noise.

\begin{table*}[h]
\centering
\caption{The precision, recall, F-score of loss and wrong event. The experiment is conducted with random initialized ResNet18 on CIFAR10 under Inst.40\% noise. We set three thresholds to calculate F-score.}
\scalebox{0.73}{
\begin{tabular}{ll|ccc|ccc|ccc}
\toprule
Threshold &  & \multicolumn{3}{c|}{0.2} & \multicolumn{3}{c|}{0.5} & \multicolumn{3}{c}{0.8} \\
\cmidrule(lr){1-2} \cmidrule(lr){3-5} \cmidrule(lr){6-8} \cmidrule(lr){9-11}
Epoch& & 10 & 30 & 60 & 10 & 30 & 60 &10 & 30 & 60 \\
\midrule
\multirow{3}{*}{Loss} & F-score & 0.49 & 0.38 & 0.29 & 0.34 & 0.29 & 0.20 & 0.00 & 0.00 & 0.00 \\
 & Precision & 0.62 & 0.76 & 0.79 & 0.71 & 0.79 & 0.75 & 0.92 & 0.90 & 0.60 \\
 & Recall & 0.41 & 0.25 & 0.18 & 0.23 & 0.18 & 0.11 & 0.00 & 0.00 & 0.00 \\
\midrule
\multirow{3}{*}{Wrong Event} & F-score & \textbf{0.60} & \textbf{0.67} & \textbf{0.70} & \textbf{0.63} & \textbf{0.70} & \textbf{0.72} & \textbf{0.70} & \textbf{0.79} & \textbf{0.76} \\
 & Precision & 0.43 & 0.51 & 0.54 & 0.46 & 0.54 & 0.56 & 0.63 & 0.88 & 0.93 \\
 & Recall & 0.99 & 0.99 & 0.99 & 0.99 & 0.99 & 0.99 & 0.77 & 0.71 & 0.65 \\
\bottomrule
\end{tabular}
}
\label{scr-f}
\end{table*}

\begin{table*}[h]
\centering
\caption{The precision, recall, F-score of loss and wrong event. The experiment is conducted with pre-trained ResNet-50 on CIFAR100 under Inst.40\% noise. We set three thresholds to calculate F-score.}
\scalebox{0.73}{
\begin{tabular}{ll|ccc|ccc|ccc}
\toprule
Threshold &  & \multicolumn{3}{c|}{0.2} & \multicolumn{3}{c|}{0.5} & \multicolumn{3}{c}{0.8} \\
\cmidrule(lr){1-2} \cmidrule(lr){3-5} \cmidrule(lr){6-8} \cmidrule(lr){9-11}
Epoch& & 5 & 20 & 30 & 5 & 20 & 30 & 5 & 20 & 30 \\
\midrule
\multirow{3}{*}{Loss} & F-score & \textbf{0.77} & 0.84 & 0.56 & 0.83 & 0.61 & 0.04 & 0.05 & 0.01 & 0.00 \\
 & Precision & 0.63 & 0.75 & 0.41 & 0.81 & 0.97 & 0.94 & 1.00 & 1.00 & 1.00 \\
 & Recall & 1.00 & 0.96 & 0.87 & 0.85 & 0.45 & 0.02 & 0.02 & 0.00 & 0.00 \\
\midrule
\multirow{3}{*}{Wrong Event} & F-score &  0.76 & \textbf{0.85} & \textbf{0.84} & \textbf{0.85} & \textbf{0.90} & \textbf{0.92} & \textbf{0.87} & \textbf{0.92} & \textbf{0.88} \\
 & Precision & 0.62 & 0.76 & 0.74 & 0.73 & 0.83 & 0.89 & 0.83 & 0.93 & 0.97 \\
 & Recall & 1.00 & 0.97 & 0.99 & 0.99 & 0.98 & 0.96 & 0.93 & 0.92 & 0.81 \\
\bottomrule
\end{tabular}
}
\label{pre-f}
\end{table*}

\begin{table*}[h!]
\centering
\caption{AUC value of wrong event and loss with random initialized ResNet-18 under different noise types and levels on CIFAR-10 (left) and CIFAR-100 (right). Higher AUC value indicates better clean sample selection ability.}
\begin{minipage}[t]{0.5\textwidth}
\centering
\scalebox{0.58}{
\begin{tabular}{lccc|ccc|ccc}
\toprule
Noise & \multicolumn{3}{c}{Sym. 60\%} & \multicolumn{3}{c}{Asym. 40\%} & \multicolumn{3}{c}{Inst. 40\%} \\
\cmidrule(lr){1-1} \cmidrule(lr){2-4} \cmidrule(lr){5-7} \cmidrule(lr){8-10}
Epoch & 20 & 60 & 100 & 20 & 60 & 100 & 20 & 60 & 100 \\
\midrule
Loss & 0.95 & 0.80 & 0.63 & 0.80 & 0.64 & 0.58 & 0.84 & 0.65 & 0.59 \\
Wrong Event & \textbf{0.96} & \textbf{0.98} & \textbf{0.97} & \textbf{0.94} & \textbf{0.94} & \textbf{0.92} & \textbf{0.94} & \textbf{0.95} & \textbf{0.94} \\
\bottomrule
\end{tabular}
}
\end{minipage}%
\hfill
\begin{minipage}[t]{0.5\textwidth}
\centering
\scalebox{0.58}{
\begin{tabular}{lccc|ccc|ccc}
\toprule
Noise & \multicolumn{3}{c}{Sym. 60\%} & \multicolumn{3}{c}{Asym. 40\%} & \multicolumn{3}{c}{Inst. 40\%} \\
\cmidrule(lr){1-1} \cmidrule(lr){2-4} \cmidrule(lr){5-7} \cmidrule(lr){8-10}
Epoch & 20 & 60 & 100 & 20 & 60 & 100 & 20 & 60 & 100 \\
\midrule
Loss & 0.85 & 0.69 & 0.60 & 0.62 & 0.56 & 0.55 & 0.79 & 0.61 & 0.59 \\
Wrong Event & \textbf{0.87} & \textbf{0.92} & \textbf{0.91} & \textbf{0.72} & \textbf{0.75} & \textbf{0.74} & \textbf{0.86} & \textbf{0.88} & \textbf{0.88} \\
\bottomrule
\end{tabular}
}
\label{random-auc}
\end{minipage}
\end{table*}

\begin{table*}[h!]
\centering
\caption{AUC value of wrong event and loss with pre-trained ResNet-50 under different noise types and levels on CIFAR-100. Higher AUC value indicates better clean sample selection ability.}
\centering
\scalebox{0.65}{
\begin{tabular}{lccc|ccc|ccc}
\toprule
Noise & \multicolumn{3}{c}{Sym. 60\%} & \multicolumn{3}{c}{Asym. 40\%} & \multicolumn{3}{c}{Inst. 40\%} \\
\cmidrule(lr){1-1} \cmidrule(lr){2-4} \cmidrule(lr){5-7} \cmidrule(lr){8-10}
Epoch & 5 & 15 & 25 & 5 & 15 & 25 & 5 & 15 & 25 \\
\midrule
Loss & 0.96 & 0.88 & 0.74 & 0.66 & 0.63 & 0.59 & 0.84 & 0.75 & 0.67 \\
Wrong Event & \textbf{0.97} & \textbf{0.98} & \textbf{0.98} & \textbf{0.76} & \textbf{0.80} & \textbf{0.79} & \textbf{0.93} & \textbf{0.95} & \textbf{0.94} \\
\bottomrule
\end{tabular}
}
\label{pre-auc}
\end{table*}

%% file: appd.tex
\section{Training time analysis}\label{appd}
We analyze the training time of IDO to understand its efficiency. Results in \autoref{time} show that IDO is faster than all baselines. For IDO itself, the running time $\approx$ BMM fitting + forward propogation. In \autoref{compute}, we compare the total training time of IDO on CIFAR-100 Inst.40\% noise with several state-of-the-art methods which leverage probabilistic model, using a single Nvidia A100 80GB GPU. 

IDO is faster than M-correction and DivideMix which involve multiple forward propagation and fitting iterations. In \autoref{compute}, we also break down the computation time for each operation in IDO. Results show that BMM is efficient, only 4.6s per epoch, which is 3.8\% of the total training time. Moreover, results show that wrong event-based BMM in IDO is faster than loss-based BMM/GMM in M-correction and DivideMix. The reason is that the robustness of wrong event helps to reduce the fitting iteration rounds and require no reinitialization to save time. While loss often varies significantly in successive epochs, needing more fitting rounds and reinitialization in each epoch. In the open-sourced code, IDO only iterates few rounds based on the last BMM (1 iteration per round, no reinitialization), while M-correction and DivideMix need many rounds and reinitialization (10 GMM iterations in DivideMix, 20 BMM iterations in M-correction) to handle loss mutation. 

\begin{table*}[h!]
\centering
\caption{Comparison of methods using probabilistic model on CIFAR-100 with Inst. 40\% Noise}
\scalebox{0.8}{ 
\begin{tabular}{l|l|l|l}
\toprule
\textbf{Method} & \textbf{Accuracy(\%)} & \textbf{Per Epoch Time(s)} & \textbf{Per Epoch Fitting Time(s)}  \\
\midrule
M-Correction(BMM) & 78.9 & 399 & 14.7  \\
DivideMix(GMM) & 81.3 & 457 & 19.5  \\
IDO(BMM) & \textbf{83.7} & \textbf{121} & \textbf{4.6}  \\
\bottomrule
\end{tabular}
}
\label{compute}
\end{table*}

%% file: appe.tex
\section{More exploratory experiments}\label{appe}
In this section, we conduct more exploratory experiments, hoping to offer more insight and inspiration to the readers and community.

\subsection{Base model vs. initial model}
After Stage 1, we already have information about clean, noisy and hard samples. It is interesting to think that it is better to train the model from the base model or directly from the initial model. Since the base model has already been trained on corrupted samples, retraining from the initial model might lead to better performance.

\begin{table*}[h!]
\centering
\caption{Comparison of initial model and base model on CIFAR-100 with different noise types}
\scalebox{0.8}{ 
\begin{tabular}{l|c|c|c}
\toprule
\textbf{Start Model} & \textbf{Sym. 60\%} & \textbf{Asym. 40\%} & \textbf{Inst. 40\%} \\
\midrule
Initial Model & 80.1 & 77.1 & 83.3  \\
Base Model & \textbf{81.5} & \textbf{78.0} & \textbf{83.7}  \\
\bottomrule
\end{tabular}
}
\label{modelchoose}
\end{table*}

We run experiments to verify this. Results are in \autoref{modelchoose}. Overall, base model yields better results. Its advantage is high when handling Sym. noise as symmetric noise is harder to fit and easy to detect. The difference is smaller when handling Asym./Inst. noise as the model fits these noise more easily, decreasing the performance. Within the relatively limited training time, using the base model can more quickly achieve better performance.

\subsection{Wrong event vs. loss accumulation}
As the single epoch loss is easy to fit noise, it is interesting to think whether accumulated loss can prevent fitting noise. We have conducted new experiments to compare accumulated loss and wrong event. We get two observations from the results in \autoref{accumulate}.

\begin{table*}[h!]
\centering
\caption{Comparison of wrong event and accumulated loss on CIFAR-100 with various noise types}
\scalebox{0.8}{ 
\begin{tabular}{l|c|c|c}
\toprule
\textbf{Noise} & \textbf{Sym. 60\%} & \textbf{Asym. 40\%} & \textbf{Inst. 40\%}  \\
\midrule
Single Loss + BMM & 77.1 & 71.9 & 72.6 \\
Accumulated Loss + GMM & 79.2 & 74.5 & 77.9  \\
Accumulated Loss + BMM & 80.1 & 75.9 & 79.1  \\
Wrong event + BMM & \textbf{81.2} & \textbf{78.3} & \textbf{83.4}  \\
\bottomrule
\end{tabular}
}
\label{accumulate}
\end{table*}

1. Accumulated loss performs better than single loss because accumulation prevents fitting noise, but the variance increases linearly when summing i.i.d. data. Besides, accidental sample forgetting can have a significant impact on the accumulated loss, increasing the variance. Thus, the distribution of loss sum is flatter than that of wrong event, which blurs the classification boundaries, making more data misclassified which drops the performance.

2. BMM performs better than GMM. In \cite{DivideMix07}, the authors point out that GMM does not fit well on Asym. noise. Therefore, BMM, which shows stronger fitting ability, performs better.

\subsection{Synthetic dataset vs. real world dataset}
We find that IDO and all baselines achieve a smaller improvement on real-world datasets compared to synthetic datasets as shown in \autoref{CombinedResults} and \autoref{realworld}, which is worth further discussion. 

We first analyze the difference between real-world noise and synthetic noise. \cite{TURN08, beyond56} pointed out that real world noise usually lies closer to the decision boundary than synthetic noise, which is harder to detect. In other words, it is easier to fit as boundary cases, slightly changing the decision boundary and thus results in less degradation of model performance than synthetic noise which will significantly change the decision boundary after fitted. As the degradation become smaller, the improvement space from w/ label noise to w/o label noise for algorithms is smaller.

\subsection{Strong algorithm collapse in CIFAR-10}\label{appe4}

The experiment result for CIFAR-10 is in \autoref{CIFAR10}. IDO still outperforms baselines. It is surprising to see a performance drop outside of UNICON, a competitive algorithm in other datasets, showing that the variation of the hyperparameters across datasets and models easily decrease the model performance. This indicates high tuning cost and performance drop for hyperparameter-based methods in real world scenarios.
 
\begin{table*}[h!]
\centering
\caption{Comparison with state-of-the-art LNL algorithms in test accuracy (\%) on CIFAR-10.}
\scalebox{0.8}{ 
\begin{tabular}{l|l|c|c|c|c|c}
\toprule
\textbf{Noise} & \textbf{Architecture} & \textbf{Sym. 20\%} & \textbf{Sym. 40\%} & \textbf{Sym. 60\%} & \textbf{Asym. 40\%} & \textbf{Inst. 40\%}  \\
\midrule
Standard & ResNet-50 & 93.2\% & 92.3\% & 88.2\% & 91.1\% & 90.9\%  \\
UNICON & ResNet-50 & 94.8\% & 93.2\% & 92.5\% & 93.5\% & 93.9\%  \\
ELR & ResNet-50 & 96.5\% & 95.8\% & 95.1\% & 95.0\% & 94.8\%  \\
DeFT & CLIP-ResNet-50 & 96.9\% & 96.6\% & 95.7\% & 93.8\% & 95.1\%  \\
DivideMix & ResNet-50 & \underline{97.1}\% & \underline{96.8\%} & \underline{96.3}\% & 93.1\% & 96.0\%  \\
DISC & ResNet-50 & 96.8\% & 96.5\% & 95.5\% & \underline{95.2}\% & \underline{96.5}\%  \\
\rowcolor{gray!40}
IDO & ResNet-50 & \textbf{97.3\%} & \textbf{96.9\%} & \textbf{96.5\%} & \textbf{95.3\%} &\textbf{96.6\%}  \\
\bottomrule
\end{tabular}
}
\label{CIFAR10}
\end{table*}

\subsection{Including more baselines}

We include InstanceGM \cite{InstanceGM57} and NAL \cite{NAL58} as our two more baselines in 3 new experiments on CIFAR-100 (see Table \ref{more baselines}).

\begin{table*}[h!]
\centering
\caption{Comparison of more baselines on CIFAR-100 with various noise types}
\scalebox{0.8}{ 
\begin{tabular}{l|c|c|c}
\toprule
\textbf{Method} & \textbf{Sym. 60\%} & \textbf{Asym. 40\%} & \textbf{Inst. 40\%}  \\
\midrule
Standard& 41.1 & 53.4 & 58.8 \\
InstanceGM& 80.5 & 76.3 & 83.1  \\
NAL& 80.9 & 77.6 & 81.3  \\
IDO& \textbf{81.4} & \textbf{78.2} & \textbf{83.8}  \\
\bottomrule
\end{tabular}
}
\label{more baselines}
\end{table*}
The results show that while InstanceGM is strong in its target setting (instance noise), it underperforms in symmetric and asymmetric noise because InstanceGM is too cautious in judging noise which leads to insufficient label correction. NAL performs well but is slightly behind IDO, particularly in the most challenging instance-dependent scenario. IDO demonstrates consistently state-of-the-art performance across all three settings.

\subsection{Experiments on broader noise levels}
To thoroughly evaluate IDO's robustness and practical applicability, we agree that testing on a wider range of noise levels is crucial. We have conducted 6 new experiments on CIFAR-100 using the pretrained ResNet-50 with more noise levels (See Table \ref{broader level}). For symmetric noise, a high noise ratio of 80\% is added. For asymmetric noise, 20\% and 45\% are added. For instance-dependent noise, 20\% and 60\% are added. Besides, a noise-free (0\%) baseline is included.

\begin{table*}[htbp!]
\centering
\caption{Performance comparison of broader noise levels}
\label{broader level}
\begin{tabular}{lcccccc}
\toprule
Method & Sym. 0\% & Sym. 80\% & Asym. 20\% & Asym. 45\% & Inst. 20\% & Inst. 60\% \\
\midrule
Standard & 82.8\% & 35.2\% & 68.5\% & 46.2\% & 70.4\% & 31.6\% \\
UNICON & 84.8\% & \textbf{64.0\%} & 83.5\% & 73.2\% & 84.9\% & 58.2\% \\
DeFT & 85.1\% & 57.6\% & 78.9\% & 65.2\% & 82.4\% & 56.5\% \\
\textbf{IDO} & \textbf{85.8\%} & 62.5\% & \textbf{85.4\%} & \textbf{74.1\%} & \textbf{85.1\%} & \textbf{60.8\%} \\
\bottomrule
\end{tabular}
\end{table*}

These results demonstrate that IDO maintains a significant performance advantage across a wide spectrum of noise types and levels, including the challenging 45\% asymmetric noise and 60\% instance-dependent noise scenario. This confirms the robustness of our approach, even under conditions far more extreme than the 8\%–38.5\% noise typically found in real-world datasets \cite{survey55}. Impressively, IDO also achieves the best performance in the noise-free (0\%) setting, showing it does not degrade performance on clean data.

As we discussed in our limitations, performance degradation is only observed at an extremely high noise ratio of 80\%, which is expected due to the severe imbalance in the wrong event distribution that affects the BMM and model convergence. We believe these new results comprehensively address the concern about robustness and practical applicability.

%% file: neurips_2025.bbl
\begin{thebibliography}{10}
\providecommand{\url}[1]{#1}
\csname url@samestyle\endcsname
\providecommand{\newblock}{\relax}
\providecommand{\bibinfo}[2]{#2}
\providecommand{\BIBentrySTDinterwordspacing}{\spaceskip=0pt\relax}
\providecommand{\BIBentryALTinterwordstretchfactor}{4}
\providecommand{\BIBentryALTinterwordspacing}{\spaceskip=\fontdimen2\font plus
\BIBentryALTinterwordstretchfactor\fontdimen3\font minus \fontdimen4\font\relax}
\providecommand{\BIBforeignlanguage}[2]{{%
\expandafter\ifx\csname l@#1\endcsname\relax
\typeout{** WARNING: IEEEtran.bst: No hyphenation pattern has been}%
\typeout{** loaded for the language `#1'. Using the pattern for}%
\typeout{** the default language instead.}%
\else
\language=\csname l@#1\endcsname
\fi
#2}}
\providecommand{\BIBdecl}{\relax}
\BIBdecl

\bibitem{Webvision01}
W.~Li, L.~Wang, W.~Li, E.~Agustsson, and L.~V. Gool, ``Webvision database: Visual learning and understanding from web data,'' \emph{arXiv preprint arXiv:1708.02862}, 2017.

\bibitem{Yan02}
Y.~Yan, R.~Rosales, G.~Fung, R.~Subramanian, and J.~Dy, ``Learning from multiple annotators with varying expertise,'' \emph{Machine Learning}, vol.~95, no.~3, pp. 291--327, 2014.

\bibitem{BenitezQuiroz03}
C.~F. Benitez-Quiroz, R.~Srinivasan, and A.~M. Martinez, ``Emotionet: An accurate, real-time algorithm for the automatic annotation of a million facial expressions in the wild,'' in \emph{Proc. CVPR}, 2016, pp. 5562--5570.

\bibitem{zhang00}
C.~Zhang, S.~Bengio, M.~Hardt, B.~Recht, and O.~Vinyals, ``Understanding deep learning requires rethinking generalization,'' in \emph{International Conference on Learning Representations}, 2017.

\bibitem{MemoryEffect10}
D.~Arpit, S.~Jastrzębski, N.~Ballas, D.~Krueger, E.~Bengio, M.~S. Kanwal, T.~Maharaj, A.~Fischer, A.~Courville, and Y.~B. et~al., ``A closer look at memorization in deep networks,'' in \emph{Proc. ICML}, 2017, pp. 233--242.

\bibitem{Coteaching04}
B.~Han, Q.~Yao, X.~Yu, G.~Niu, M.~Xu, W.~Hu, I.~Tsang, and M.~Sugiyama, ``Co-teaching: Robust training of deep neural networks with extremely noisy labels,'' in \emph{Proc. NeurIPS}, vol.~31, 2018.

\bibitem{O2U11}
J.~Huang, L.~Qu, R.~Jia, and B.~Zhao, ``O2u-net: A simple noisy label detection approach for deep neural networks,'' in \emph{2019 IEEE/CVF International Conference on Computer Vision (ICCV)}, 2019, pp. 3325--3333.

\bibitem{ELR05}
S.~Liu, J.~Niles-Weed, N.~Razavian, and C.~Fernandez-Granda, ``Early-learning regularization prevents memorization of noisy labels,'' in \emph{Proc. NeurIPS}, vol.~33, 2020, pp. 20\,331--20\,342.

\bibitem{Triple20}
X.~Liang, L.~Yao, X.~Liu, and Y.~Zhou, ``Tripartite: Tackle noisy labels by a more precise partition,'' \emph{ArXiv}, vol. abs/2202.09579, 2022.

\bibitem{APL43}
X.~Ye, X.~Li, S.~Dai, T.~Liu, Y.~Sun, and W.~Tong, ``Active negative loss functions for learning with noisy labels,'' in \emph{Thirty-seventh Conference on Neural Information Processing Systems}, 2023.

\bibitem{DISC14}
Y.~Li, H.~Han, S.~Shan, and X.~Chen, ``Disc: Learning from noisy labels via dynamic instance-specific selection and correction,'' in \emph{Proceedings of the IEEE/CVF Conference on Computer Vision and Pattern Recognition}, 2023, pp. 24\,070--24\,079.

\bibitem{TURN08}
S.~Ahn, S.~Kim, J.~Ko, and S.-Y. Yun, ``Fine-tuning pre-trained models for robustness under noisy labels,'' in \emph{Proceedings of the Thirty-Third International Joint Conference on Artificial Intelligence, {IJCAI-24}}, 8 2024, pp. 3643--3651.

\bibitem{DEFT09}
T.~Wei, H.-T. Li, C.-S. Li, J.-X. Shi, Y.-F. Li, and M.-L. Zhang, ``Vision-language models are strong noisy label detectors,'' in \emph{The Thirty-eighth Annual Conference on Neural Information Processing Systems}, 2024.

\bibitem{CLIPCleaner53}
C.~Feng, G.~Tzimiropoulos, and I.~Patras, ``Clipcleaner: Cleaning noisy labels with clip,'' in \emph{Proceedings of the 32nd ACM International Conference on Multimedia}.\hskip 1em plus 0.5em minus 0.4em\relax New York, NY, USA: Association for Computing Machinery, 2024, p. 876–885.

\bibitem{UNICON33}
N.~Karim, M.~N. Rizve, N.~Rahnavard, A.~Mian, and M.~Shah, ``Unicon: Combating label noise through uniform selection and contrastive learning,'' in \emph{2022 IEEE/CVF Conference on Computer Vision and Pattern Recognition (CVPR)}, 2022, pp. 9666--9676.

\bibitem{DivideMix07}
J.~Li, R.~Socher, and S.~C.~H. Hoi, ``Dividemix: Learning with noisy labels as semi-supervised learning,'' in \emph{Proc. ICLR}, 2019.

\bibitem{RoCL12}
T.~Zhou, S.~Wang, and J.~Bilmes, ``Robust curriculum learning: from clean label detection to noisy label self-correction,'' in \emph{International Conference on Learning Representations}, 2021.

\bibitem{forgettingevent21}
M.~Toneva, A.~Sordoni, R.~T. des Combes, A.~Trischler, Y.~Bengio, and G.~J. Gordon, ``An empirical study of example forgetting during deep neural network learning,'' in \emph{International Conference on Learning Representations}, 2019.

\bibitem{Latestop22}
S.~Yuan, L.~Feng, and T.~Liu, ``Late stopping: Avoiding confidently learning from mislabeled examples,'' \emph{2023 IEEE/CVF International Conference on Computer Vision (ICCV)}, pp. 16\,033--16\,042, 2023.

\bibitem{soseleto37}
O.~Litany and D.~Freedman, ``{SOSELETO}: A unified approach to transfer learning and training with noisy labels,'' 2019.

\bibitem{MLC36}
G.~Zheng, A.~H. Awadallah, and S.~T. Dumais, ``Meta label correction for noisy label learning,'' in \emph{AAAI Conference on Artificial Intelligence}, 2021.

\bibitem{rankmatch38}
Z.~Zhang, W.~Chen, C.~Fang, Z.~Li, L.~Chen, L.~Lin, and G.~Li, ``Rankmatch: Fostering confidence and consistency in learning with noisy labels,'' in \emph{2023 IEEE/CVF International Conference on Computer Vision (ICCV)}, 2023, pp. 1644--1654.

\bibitem{co-learning39}
C.~Tan, J.~Xia, L.~Wu, and S.~Z. Li, ``Co-learning: Learning from noisy labels with self-supervision,'' in \emph{Proceedings of the 29th ACM International Conference on Multimedia}, 2021, p. 1405–1413.

\bibitem{M-correction16}
E.~Arazo, D.~Ortego, P.~Albert, N.~O'Connor, and K.~Mcguinness, ``Unsupervised label noise modeling and loss correction,'' in \emph{Proceedings of the 36th International Conference on Machine Learning}, ser. Proceedings of Machine Learning Research, vol.~97, 2019, pp. 312--321.

\bibitem{vit31}
A.~Dosovitskiy, L.~Beyer, A.~Kolesnikov, D.~Weissenborn, X.~Zhai, T.~Unterthiner, M.~Dehghani, M.~Minderer, G.~Heigold, S.~Gelly, J.~Uszkoreit, and N.~Houlsby, ``An image is worth 16x16 words: Transformers for image recognition at scale,'' in \emph{International Conference on Learning Representations}, 2021.

\bibitem{CLIP49}
A.~Radford, J.~W. Kim, C.~Hallacy, A.~Ramesh, G.~Goh, S.~Agarwal, G.~Sastry, A.~Askell, P.~Mishkin, J.~Clark, G.~Krueger, and I.~Sutskever, ``Learning transferable visual models from natural language supervision,'' in \emph{International Conference on Machine Learning}, 2021.

\bibitem{C2D29}
E.~Zheltonozhskii, C.~Baskin, A.~Mendelson, A.~M. Bronstein, and O.~Litany, ``{ Contrast to Divide: Self-Supervised Pre-Training for Learning with Noisy Labels },'' in \emph{2022 IEEE/CVF Winter Conference on Applications of Computer Vision (WACV)}, 2022, pp. 387--397.

\bibitem{SCL52}
S.~Li, X.~Xia, S.~Ge, and T.~Liu, ``Selective-supervised contrastive learning with noisy labels,'' \emph{2022 IEEE/CVF Conference on Computer Vision and Pattern Recognition (CVPR)}, pp. 316--325, 2022.

\bibitem{SIMCLR51}
T.~Chen, S.~Kornblith, M.~Norouzi, and G.~Hinton, ``A simple framework for contrastive learning of visual representations,'' in \emph{Proceedings of the 37th International Conference on Machine Learning}.\hskip 1em plus 0.5em minus 0.4em\relax JMLR.org, 2020.

\bibitem{MoCO50}
K.~He, H.~Fan, Y.~Wu, S.~Xie, and R.~Girshick, ``Momentum contrast for unsupervised visual representation learning,'' in \emph{2020 IEEE/CVF Conference on Computer Vision and Pattern Recognition (CVPR)}, 2020, pp. 9726--9735.

\bibitem{resnet32}
K.~He, X.~Zhang, S.~Ren, and J.~Sun, ``Deep residual learning for image recognition,'' in \emph{2016 IEEE Conference on Computer Vision and Pattern Recognition (CVPR)}, 2016, pp. 770--778.

\bibitem{Convnext47}
Z.~Liu, H.~Mao, C.-Y. Wu, C.~Feichtenhofer, T.~Darrell, and S.~Xie, ``A convnet for the 2020s,'' in \emph{2022 IEEE/CVF Conference on Computer Vision and Pattern Recognition (CVPR)}, 2022, pp. 11\,966--11\,976.

\bibitem{FocalLoss45}
T.-Y. Lin, P.~Goyal, R.~Girshick, K.~He, and P.~Dollár, ``Focal loss for dense object detection,'' in \emph{2017 IEEE International Conference on Computer Vision (ICCV)}, 2017, pp. 2999--3007.

\bibitem{GHM46}
B.~Li, Y.~Liu, and X.~Wang, ``Gradient harmonized single-stage detector,'' in \emph{Proceedings of the Thirty-Third AAAI Conference on Artificial Intelligence and Thirty-First Innovative Applications of Artificial Intelligence Conference and Ninth AAAI Symposium on Educational Advances in Artificial Intelligence}, ser. AAAI'19/IAAI'19/EAAI'19.\hskip 1em plus 0.5em minus 0.4em\relax AAAI Press, 2019.

\bibitem{CL44}
Y.~Bengio, J.~Louradour, R.~Collobert, and J.~Weston, ``Curriculum learning,'' in \emph{Proceedings of the 26th Annual International Conference on Machine Learning}, ser. ICML '09, 2009, p. 41–48.

\bibitem{bmm40}
Z.~Ma and A.~Leijon, ``Bayesian estimation of beta mixture models with variational inference,'' \emph{IEEE Transactions on Pattern Analysis and Machine Intelligence}, vol.~33, no.~11, pp. 2160--2173, 2011.

\bibitem{LabelWave15}
S.~Yuan, L.~Feng, and T.~Liu, ``Early stopping against label noise without validation data,'' in \emph{The Twelfth International Conference on Learning Representations}, 2024.

\bibitem{AugDisc18}
K.~Nishi, Y.~Ding, A.~Rich, and T.~Hollerer, ``Augmentation strategies for learning with noisy labels,'' in \emph{Proceedings of the IEEE/CVF Conference on Computer Vision and Pattern Recognition (CVPR)}, June 2021, pp. 8022--8031.

\bibitem{RRL17}
J.~Li, C.~Xiong, and S.~C. Hoi, ``Learning from noisy data with robust representation learning,'' in \emph{2021 IEEE/CVF International Conference on Computer Vision (ICCV)}, 2021, pp. 9465--9474.

\bibitem{cifar23}
A.~Krizhevsky, ``Learning multiple layers of features from tiny images,'' 2009.

\bibitem{tinyimagenet24}
J.~Wu, Q.~Zhang, and G.~Xu, ``Tiny imagenet challenge,'' Technical report, 2017.

\bibitem{IDN25}
X.~Xia, T.~Liu, B.~Han, N.~Wang, M.~Gong, H.~Liu, G.~Niu, D.~Tao, and M.~Sugiyama, ``Part-dependent label noise: Towards instance-dependent label noise,'' \emph{Advances in Neural Information Processing Systems}, vol.~33, pp. 7597--7610, 2020.

\bibitem{CIFAR100N26}
J.~Wei, Z.~Zhu, H.~Cheng, T.~Liu, G.~Niu, and Y.~Liu, ``Learning with noisy labels revisited: A study using real-world human annotations,'' in \emph{International Conference on Learning Representations}, 2022.

\bibitem{Clothing1m27}
T.~Xiao, T.~Xia, Y.~Yang, C.~Huang, and X.~Wang, ``Learning from massive noisy labeled data for image classification,'' in \emph{Proceedings of the IEEE Conference on Computer Vision and Pattern Recognition (CVPR)}, 2015.

\bibitem{efficient30}
J.~Ko, S.~Ahn, and S.-Y. Yun, ``{EFFICIENT} {UTILIZATION} {OF} {PRE}-{TRAINED} {MODEL} {FOR} {LEARNING} {WITH} {NOISY} {LABELS},'' in \emph{ICLR 2023 Workshop on Pitfalls of limited data and computation for Trustworthy ML}, 2023.

\bibitem{SCE41}
Y.~Wang, X.~Ma, Z.~Chen, Y.~Luo, J.~Yi, and J.~Bailey, ``Symmetric cross entropy for robust learning with noisy labels,'' in \emph{2019 IEEE/CVF International Conference on Computer Vision (ICCV)}, 2019, pp. 322--330.

\bibitem{LongReMix34}
F.~R. Cordeiro, R.~Sachdeva, V.~Belagiannis, I.~Reid, and G.~Carneiro, ``Longremix: Robust learning with high confidence samples in a noisy label environment,'' \emph{Pattern Recognition}, vol. 133, p. 109013, 2023.

\bibitem{ProMix35}
R.~Xiao, Y.~Dong, H.~Wang, L.~Feng, R.~Wu, G.~Chen, and J.~Zhao, ``Promix: Combating label noise via maximizing clean sample utility,'' in \emph{Proceedings of the Thirty-Second International Joint Conference on Artificial Intelligence, {IJCAI-23}}, E.~Elkind, Ed., 8 2023, pp. 4442--4450.

\bibitem{survey55}
H.~Song, M.~Kim, D.~Park, and J.~Lee, ``Learning from noisy labels with deep neural networks: {A} survey,'' \emph{CoRR}, 2020.

\bibitem{mixup48}
H.~Zhang, M.~Cisse, Y.~N. Dauphin, and D.~Lopez-Paz, ``mixup: Beyond empirical risk minimization,'' in \emph{International Conference on Learning Representations}, 2018.

\bibitem{fluctuateevent54}
Q.~Wei, H.~Sun, X.~Lu, and Y.~Yin, ``Self-filtering: A noise-aware sample selection for label noise with confidence penalization,'' in \emph{ECCV}, 2022, pp. 516--532.

\bibitem{beyond56}
L.~Jiang, D.~Huang, M.~Liu, and W.~Yang, ``Beyond synthetic noise: Deep learning on controlled noisy labels,'' in \emph{Proceedings of the 37th International Conference on Machine Learning}, 2020, pp. 4804--4815.

\bibitem{InstanceGM57}
\BIBentryALTinterwordspacing
A.~Garg, C.~Nguyen, R.~Felix, T.-T. Do, and G.~Carneiro, ``Instance-dependent noisy label learning via graphical modelling,'' 2022. [Online]. Available: \url{https://arxiv.org/abs/2209.00906}
\BIBentrySTDinterwordspacing

\bibitem{NAL58}
\BIBentryALTinterwordspacing
Y.~Lu, Y.~Bo, and W.~He, ``Noise attention learning: Enhancing noise robustness by gradient scaling,'' in \emph{Advances in Neural Information Processing Systems}, S.~Koyejo, S.~Mohamed, A.~Agarwal, D.~Belgrave, K.~Cho, and A.~Oh, Eds., vol.~35.\hskip 1em plus 0.5em minus 0.4em\relax Curran Associates, Inc., 2022, pp. 23\,164--23\,177. [Online]. Available: \url{https://proceedings.neurips.cc/paper_files/paper/2022/file/92864e1191ed272deb0914b3bb50f97c-Paper-Conference.pdf}
\BIBentrySTDinterwordspacing

\end{thebibliography}
